\documentclass[10pt,journal,compsoc]{IEEEtran}
\usepackage{color}
\usepackage{xcolor}
\usepackage{graphicx}
\usepackage{colortbl}
\usepackage{subfigure}
\usepackage{amsmath,amssymb} 
\usepackage{multirow}
\usepackage{makecell}
\usepackage{bbding}
\usepackage{tabularx}
\usepackage{booktabs}
\usepackage[pagebackref=true,breaklinks=true,colorlinks,bookmarks=false]{hyperref}
\usepackage{bm}
\usepackage{url}
\usepackage{tikz}
\usetikzlibrary{spy}

\begin{document}
\title{SpectralGPT: Spectral Remote Sensing Foundation Model}

\author{Danfeng~Hong,~\IEEEmembership{Senior Member,~IEEE,}
        Bing~Zhang,~\IEEEmembership{Fellow,~IEEE,}
        Xuyang~Li,
        Yuxuan~Li,
        Chenyu~Li,
        Jing~Yao,~\IEEEmembership{Member,~IEEE,}
        Naoto~Yokoya,~\IEEEmembership{Member,~IEEE,}
        Hao~Li,~\IEEEmembership{Member,~IEEE,}
        Pedram~Ghamisi,~\IEEEmembership{Senior Member,~IEEE,}
        Xiuping~Jia,~\IEEEmembership{Fellow,~IEEE,}
        Antonio~Plaza,~\IEEEmembership{Fellow,~IEEE,}
        Paolo~Gamba,~\IEEEmembership{Fellow,~IEEE,}
        Jon~Atli~Benediktsson,~\IEEEmembership{Fellow,~IEEE,}
        Jocelyn~Chanussot,~\IEEEmembership{Fellow,~IEEE}        
\IEEEcompsocitemizethanks{
\vspace{-5pt}
\IEEEcompsocthanksitem D. Hong, X. Li, and Y. Li are with the Aerospace Information Research Institute, Chinese Academy of Sciences, Beijing 100094, China, and also with the School of Electronic, Electrical and Communication Engineering, University of Chinese Academy of Sciences, Beijing 100049, China.
\IEEEcompsocthanksitem B. Zhang is with the Aerospace Information Research Institute, Chinese Academy of Sciences, Beijing 100094, China, and also with the College of Resources and Environment, University of Chinese Academy of Sciences, Beijing 100049, China.
\IEEEcompsocthanksitem C. Li is with the Aerospace Information Research Institute, Chinese Academy of Sciences, Beijing 100094, China, and also with the School of Mathematics, Southeast University, Nanjing 210096, China.
\IEEEcompsocthanksitem J. Yao is with the Aerospace Information Research Institute, Chinese Academy of Sciences, Beijing 100094, China. 
\IEEEcompsocthanksitem N. Yokoya is with the Graduate School of Frontier Sciences, the University of Tokyo, Chiba 277–8561, Japan.
\IEEEcompsocthanksitem H. Li is with the Big Geospatial Data Management, Technical University of Munich, Munich 85521, Germany.
\IEEEcompsocthanksitem P. Ghamisi is with the Helmholtz-Zentrum Dresden-Rossendorf, Helmholtz Institute Freiberg for Resource Technology, Freiberg 09599, Germany, and also with the Institute of Advanced Research in Artificial Intelligence, Vienna 1030, Austria.
\IEEEcompsocthanksitem X. Jia is with the School of Engineering and Information Technology, University of New South Wales, Canberra, ACT 2612, Australia.
\IEEEcompsocthanksitem A. Plaza is with the Hyperspectral Computing Laboratory, Department of Technology of Computers and Communications, Escuela Polit\'ecnica, University of Extremadura, C\'aceres 10003, Spain.
\IEEEcompsocthanksitem P. Gamba is with the Department of Electrical, Computer and Biomedical Engineering, University of Pavia, Pavia 27100, Italy.
\IEEEcompsocthanksitem J. Benediktsson is with the Faculty of Electrical and Computer Engineering, University of Iceland, Reykjavik 102, Iceland.
\IEEEcompsocthanksitem J. Chanussot is with Univ. Grenoble Alpes, Inria, CNRS, Grenoble INP, LJK, Grenoble 38000, France.
\IEEEcompsocthanksitem B. Zhang is the corresponding author (zb@radi.ac.cn).
}}

\markboth{Accepted by IEEE TPAMI}%
{Shell \MakeLowercase{\textit{et al.}}: Bare Advanced Demo of IEEEtran.cls for IEEE Computer Society Journals}

\IEEEtitleabstractindextext{%
\begin{abstract}
The foundation model has recently garnered significant attention due to its potential to revolutionize the field of visual representation learning in a self-supervised manner. While most foundation models are tailored to effectively process RGB images for various visual tasks, there is a noticeable gap in research focused on spectral data, which offers valuable information for scene understanding, especially in remote sensing (RS) applications. To fill this gap, we created for the first time a universal RS foundation model, named SpectralGPT, which is purpose-built to handle spectral RS images using a novel 3D generative pretrained transformer (GPT). Compared to existing foundation models, SpectralGPT 1) accommodates input images with varying sizes, resolutions, time series, and regions in a progressive training fashion, enabling full utilization of extensive RS big data; 2) leverages 3D token generation for spatial-spectral coupling; 3) captures spectrally sequential patterns via multi-target reconstruction; 4) trains on one million spectral RS images, yielding models with over 600 million parameters. Our evaluation highlights significant performance improvements with pretrained SpectralGPT models, signifying substantial potential in advancing spectral RS big data applications within the field of geoscience across four downstream tasks:  single/multi-label scene classification, semantic segmentation, and change detection.
\end{abstract}

\begin{IEEEkeywords}
Artificial intelligence, deep learning, downstream, foundation model, tensor masked modeling, progressive, remote sensing, spectral data, transformer.
\end{IEEEkeywords}}

\maketitle

\IEEEdisplaynontitleabstractindextext

%
\IEEEpeerreviewmaketitle

\section{Introduction}
\label{sec:introduction}

\IEEEPARstart{S}{pectral} imaging is capable of capturing a vast array of spectral information, thereby enabling highly accurate analysis and recognition of objects and scenes beyond what is possible with RGB data alone \cite{goetz1985imaging,mei2022hyperspectral,zhou2023rethinking}. This has made multi/hyper-spectral (MS/HS) remote sensing (RS) data the preferred tool of choice and a key component in a wide range of Earth Observation (EO) applications \cite{reichstein2019deep,he2020non,hong2023decoupled,mei2023rotation}, including land use/land cover mapping, ecosystem monitoring, weather forecasting, energy resource development, biodiversity conservation, and geological exploration.

\begin{figure*}[!t]
      \centering
	   \includegraphics[width=1\textwidth]{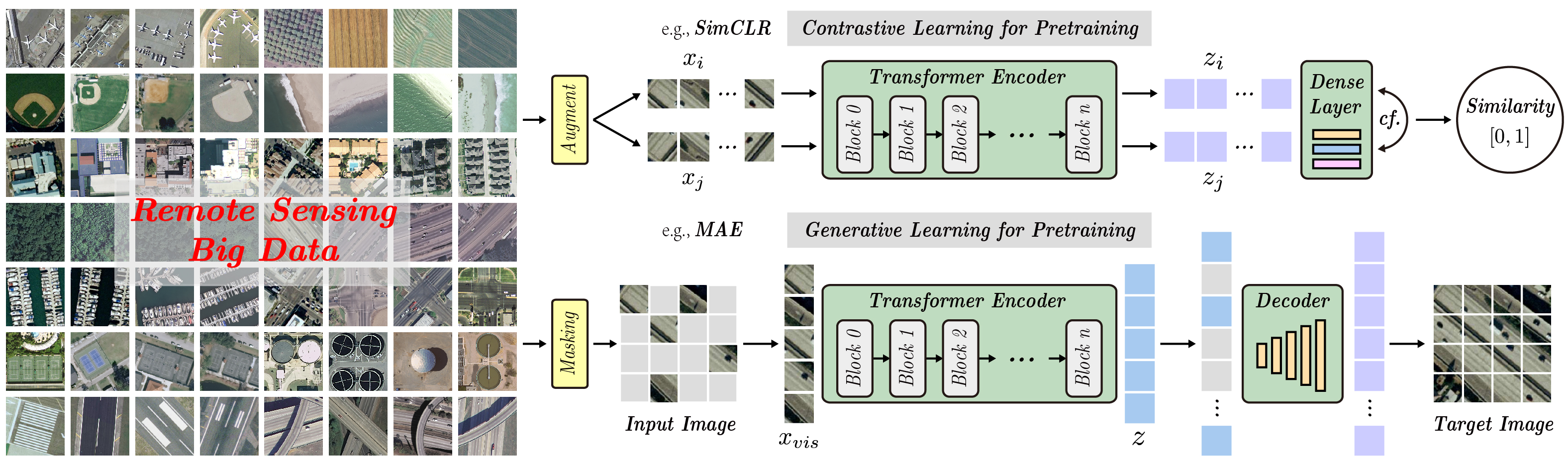}
      \caption{An illustration to clarify the differences between contrastive learning and generative learning for pretraining in terms of the RS foundation model.}
\label{fig:motivation}
\end{figure*}

The rapid expansion in the availability and accessibility of spectral data from RS satellite missions, such as Landsat-8/9, Sentinel-2, Gaofen-1/2/6, and others, has further opened up opportunities for new discoveries and advancements in fields related to EO \cite{hong2023cross}. Nevertheless, this growth also gives rise to two challenging difficulties that require prompt attention and effective solutions.
\begin{itemize}
    \item \textbf{Limited information extraction and mining capabilities from massive spectral data.} The existing expert-centric and data-driven models have reached their limits and are insufficient for effectively learning visual representations from such vast amounts of spectral RS data. There is an urgent need to create new-generation models to improve the intelligent processing and analysis capabilities of spectral RS big data to a level that matches its volume.
    \item \textbf{Limited prediction and interpretation abilities for downstream EO tasks on a few label cases.} Compared to the availability of spectral RS data, there is a scarcity of corresponding labels at both the pixel and image levels. This shortage of labeled data hinders the application of full supervision in deep learning and AI models for practical EO tasks. Urgent action is needed to create RS foundation models embodied with spectral knowledge.
\end{itemize}

The development of foundation models \cite{bommasani2021opportunities} based on pretraining techniques is currently experiencing a remarkable surge, driven by advancements in self-supervised learning techniques \cite{tian2023recent} and the transformative capabilities of transformer-based methods \cite{vaswani2017attention}. Notably, this surge is particularly evident in the domains of natural language processing and computer vision. The generative pretrained transformer (GPT), deriving out the well-known ChatGPT, is an acknowledged and representative method in the research of foundation models. These pretraining agent tasks are usually divided into contrastive learning \cite{radford2021learning} and generative learning \cite{liu2021modality}. As the name suggests, the former aims to teach the model to differentiate between similar and dissimilar examples, while the latter focuses on training a model to generate new data or recover complete data from partial observations. Their differences are illustrated in Fig. \ref{fig:motivation}. Two representative frameworks in contrastive learning are momentum contrast (MoCo) \cite{he2020momentum} and simple contrastive learning (SimCLR) \cite{chen2020simple}. MoCo introduces momentum updates to improve the contrastive learning process, while SimCLR leverages data augmentations to enhance the variety and complexity of the image pairs used for contrastive learning. There have been numerous variants of the MoCo and SimCLR frameworks developed since their initial proposals. These variants aim to address specific challenges or limitations of the original frameworks or to further improve their performance. For example, some variants of SimCLR have incorporated new types of data augmentations or improved the training objective \cite{chen2020big,caron2020unsupervised, chen2021exploring}, while some variants of MoCo have explored different momentum update strategies or used additional losses to improve contrastive learning \cite{chen2020improved, xiong2020loco, xie2021propagate}. Accompanied by the rise of vision transformers (ViT) \cite{dosovitskiy2020image}, there has been significant progress in generative learning based on masked image modeling (MIM) for visual pretraining tasks. Bidirectional encoder representation from image transformers (BEiT), as presented in \cite{bao2022beit}, is a prominent example of a MIM architecture built on top of the ViT. MIM allows for the input of all image patches, which provides flexibility for adapting to various network architectures. In \cite{du2023spectral}, a novel convolution and transformer joint network was proposed, which achieves high-accuracy spectral reconstruction in large-scale complex scenes. Yet the high computational cost associated with MIM can to some extent limit its practical use in certain applications. He \textit{et al.} \cite{he2022masked} proposed masked autoencoders (MAE) as a particular alternative to MIM. MAE has emerged as a crucial baseline architecture in the GPT family, which has shown great potential in pretraining tasks. In MAE, unmasked patches or pixels are used to reconstruct those that are masked. This approach is computationally more efficient and also enhances the inference ability of the pretrained models, thereby making it more practical for various applications. 

However, these advanced models have been relatively underexplored in RS. Wang \textit{et al.} \cite{wang2022advancing} trained a plain vision transformer with 100 million parameters on RGB images towards RS task design and developed a new rotated varied-size window attention mechanism for fine-tuning the model on downstream tasks. Unlike MAE-based methods that rely on only a few visible image patches to infer the entire image, Sun \textit{et al.} \cite{sun2022ringmo} considered all image patches, whether masked or unmasked, by implementing a MIM strategy in their RS pretraining models. Despite the increased computational cost and the reduction in inference efficiency, MIM allows for the flexible use of various deep architectures as network backbones, such as ViT and Swin transformers \cite{liu2021swin}. The success of these two initial studies indicates the significant potential of pretrained models for applications in RS. The rapid progress in imaging spectroscopy has solidified the significance of spectral RS in EO. This prominence arises from its unique ability to effectively utilize the wealth of spectral information available. However, existing RS foundation models encounter challenges when applied to spectral data due to their limited capacity to model multi-band data. The specific gaps between spectral data and existing foundation models can be summarized as follows.

\begin{itemize}
    \item \textbf{Gap 1 (\textit{cf.} existing RS foundation models):} 
    They often struggle to capture spatial-spectral representations inherent in 3D tensor data. A majority of these models are predominantly designed for processing data resembling RGB imagery, which constrains their capability to fully capture and characterize spectral information. Consequently, their applicability to handle such data types remains constrained.
    \item \textbf{Gap 2 (\textit{cf.} foundation models for video data):} There have been foundation models designed for video data in computer vision \cite{tong2022videomae, feichtenhofer2022masked}, yet there are significant differences between video data and spectral data. The main distinctions lie in the varying content between continuous frames in videos and the redundancy that often exists between all frames. As a result, the network designs pretrained for video data are often not well-suited for spectral data.
    \item \textbf{Gap 3 (\textit{cf.} foundation models for spectral data):} 
    Indeed, recent research about foundation models for spectral data has been relatively scarce. Only one conference paper, namely SatMAE \cite{cong2022satmae}, has delved into the utilization of pretrained transformers, e.g., MAE, for spectral satellite images. SatMAE's central design approach involves grouping adjacent spectral bands, akin to RGB bands. However, this practice inadvertently disrupts spectral continuity, leading to a suboptimal capture of 3D spatial-spectral coupling traits and spectrally sequential data. Moreover, constraints pertaining to the number of pretraining samples and effective training strategies have further impeded performance enhancements in this context.
\end{itemize}

\begin{figure*}[!t]
      \centering	   
      \includegraphics[width=1\textwidth]{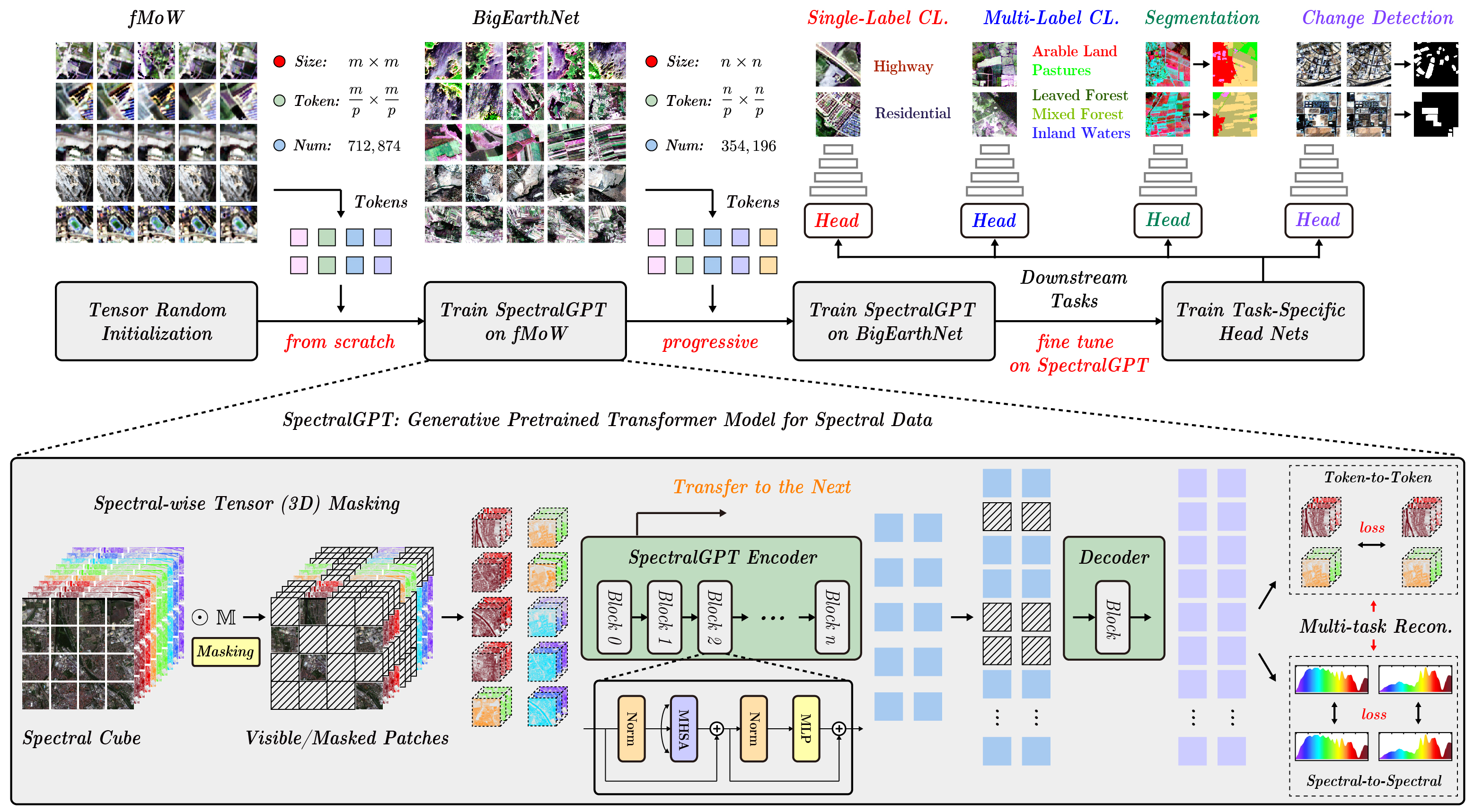}
      \caption{An illustrative workflow of the proposed SpectralGPT foundation model and its adaptation to downstream tasks. In the pretraining phase, SpectralGPT starts to train the model from scratch on one dataset (e.g., fMoW-S2, with 712,874 images) with a (3D) tensor-based random weight initialization. Subsequently, the model undergoes progressive training on more datasets (e.g., BigEarthNet-S2, with 354,196 images) with varying image sizes, time series information, and geographic regions. SpectralGPT is constructed following the MAE architecture \cite{he2022masked} and incorporates spectral-wise tensor (3D) masking, where 90\% of the tokens are masked out. For downstream tasks, such as classification, segmentation, and change detection, the pretrained SpectralGPT is connected with task-specific Head networks to be trained and then performs fine-tuning.}
\label{fig:workflow}
\end{figure*}

To fill these gaps, we devise SpectralGPT, a groundbreaking RS foundation model meticulously tailored for spectral data. SpectralGPT features pioneering elements, such as a 3D masking strategy, an encoder for learning visual representations from spatial-spectral mixed tokens, and a decoder with multi-target reconstruction for preserving spectrally sequential characteristics. These innovations significantly enhance SpectralGPT's ability to learn intrinsic knowledge representations from spectral data, providing valuable insights for scene understanding in various RS downstream applications. Fig. \ref{fig:workflow} illustrates a visual overview of SpectralGPT's pretraining and its versatile application in diverse downstream tasks, underscoring its profound contributions.

\begin{itemize}
    \item \textbf{Customized foundation model for spectral data:} SpectralGPT is the first purpose-built foundation model designed explicitly for spectral RS data. SpectralGPT considers unique characteristics of spectral data, i.e., spatial-spectral coupling and spectral sequentiality, in the MAE framework with a simple yet effective 3D GPT network. 
    \item \textbf{Large-scale training data:} SpectralGPT is trained on an extensive dataset derived from the Sentinel-2 satellite with over one million spectral images. This effort culminates in the creation of three distinct model iterations--Base, Large, and Huge--comprising approximately 100 million, 300 million, and 600 million parameters, respectively.
    \item \textbf{Flexibility in pretraining:} SpectralGPT employs a progressive training strategy, enabling it to process input images with varying sizes, resolutions, time series, and geographical regions. SpectralGPT is a comprehensive framework and a collective term, referring to the model pretrained exclusively on a single dataset in our case. Correspondingly, the model progressively pretrained on diverse datasets is denoted as SpectralGPT$^{+}$. This innovative design exposes the model's encoder to a diverse array of information, ultimately enhancing its capability to represent a wide range of features effectively.
    \item \textbf{Advanced 3D masking and reconstruction:} SpectralGPT leverages a 3D tensor-shaped spatial-spectral mask with a masking rate of at least 90\% on spectral RS data. Additionally, it employs a groundbreaking multi-target reconstruction strategy to comprehensively capture locally spatial-spectral characteristics and spectrally sequential information. These innovations substantially improve the model's learning capabilities through inference.
    \item \textbf{Superior performance across downstream tasks:} SpectralGPT's impact extends to downstream RS models, where it outperforms existing state-of-the-art (SOTA) competitors across various tasks, including single/multi-label scene classification, semantic segmentation, and change detection.
    \item \textbf{New benchmark dataset:} We curate a new benchmark dataset, named SegMunich, which focuses on urban areas and their adjacent neighborhoods within Munich City, Germany. This dataset is designed to cater to the requirements of semantic segmentation tasks with 13 classes to facilitate downstream analysis.
\end{itemize}

At the time of submission, the trained model of SpectralGPT and SpectralGPT+ has been released at \url{https://doi.org/10.5281/zenodo.10533809}.
Interested users can apply for the SpectralGPT/SpectralGPT+ encoded feature maps by emailing authors the selected image files with the required format along with a short description of the intended use. After the review, we will release an open-source implementation of the proposed model along with instructions for training/testing at the same link to further encourage public use.

\section{The Proposed SpectralGPT}
\subsection{A Brief Recall of MAE}
MAE is a simple autoencoding method \cite{vincent2010stacked} that enables the reconstruction of the original signal. Like all autoencoders, MAE includes an encoder that maps the observed signal to a potential representation and a decoder that reconstructs the original signal from the potential representation. However, in contrast to classical autoencoders, MAE uses an asymmetric design that enables the encoder to operate only on partial and observed signals (without mask tokens). Additionally, MAE employs a lightweight decoder to reconstruct the complete signal from potential representations and mask tokens.

In detail, the implementation process of the MAE can be broken down into the following steps:

{\bf Step 1.} Given the input image $\bm{x}$ with $H\times W$ pixels by $C$ dimensions, the strategy in ViT is adapted to divide it into regular, non-overlapping patches with the size of $p\times p\times C$, denoted as $\bm{x}=\{\bm{x}_{1}, \bm{x}_{2}, ..., \bm{x}_{i},...,\bm{x}_{\frac{H}{p}\times \frac{W}{p}}\}$.

{\bf Step 2.} Next, a masking operation is performed on these patches to identify visible (or unmasked) and masked patches, i.e., $\bm{x}_{vis}=\{\bm{x}_{i}|i\in vis\}$. Only the visible patches are sent into the to-be-learned encoder.

{\bf Step 3.} The encoder $f_{en}$ is implemented using ViT, where each visible patch is first linearly projected by a shared matrix $\bm{E}_{s}$, combined with positional embeddings $\bm{E}_{pos}$, and then processed through a series of transformer blocks. Thus, the encoder output in the $i$-th patch can be expressed as $\bm{z}_{i}=f_{en}(\bm{E}_{s}\bm{x}_{i}+\bm{E}_{pos})$.

{\bf Step 4.} The input to the MAE decoder, denoted by $g_{de}$, is a complete set of tokens that includes the encoded visible patches and mask tokens (e.g., $\bm{z}_{m}$). The encoded features, which are the latent representations from the encoder, and the mask tokens are used as inputs and combined with positional embeddings to the lightweight ViT decoder. The final layer of the decoder is a linear projection (e.g., $\bm{W}$) that outputs several channels equal to the number of pixels in a patch.  The output is then reshaped to reconstruct the image as $\bm{\hat{x}}=g_{de}(\bm{W}([\bm{z}_{vis},\bm{z}_{m}])+pos)$, where $\bm{z}_{vis}$ is the encoded representations of visible patches.

{\bf Step 5. } The loss function used in MAE is the mean squared error (MSE), and it is calculated for the visible and masked patches (similar to BERT \cite{devlin2018bert}), i.e., $\mathcal{L}=\frac{1}{vis+mask}\sum_{i\in vis\bigcup mask}(\bm{x}_{i}-\bm{\hat{x}}_{i})^{2}$.

It is worth noting that a normalization approach is performed, i.e., the mean and standard deviation of pixel values in each patch are calculated, and the patch is normalized accordingly, i.e., $\bm{x}_{norm}=\{\frac{\bm{x}_{i}-u_{i}}{\sigma_{i}}|i\in vis\}$. In this case, the encoder reconstruction task changes to reconstruct normalized pixel values. 

\subsection{Methodological Overview of SpectralGPT}
Our SpectralGPT model is structured with three key components: 3D masking for processing spectral data, an encoder to learn spectrally visual representations, and a decoder for multi-target reconstruction. What sets our approach apart is a progressive training manner, where the model is trained using diverse types of spectral data. This strategy enhances the proposed SpectralGPT foundation model, endowing it with greater flexibility, robustness, and generalization capabilities. Fig. \ref{fig:workflow} provides an illustrative workflow of the proposed SpectralGPT with various downstream tasks.

\subsection{3D Masking on Spectral Data}
Inspired by the spacetime-agnostic sampling in video-like data using the MAE-based framework \cite{feichtenhofer2022masked}, we model multi-band spectral images as 3D tensor data. To achieve this, we implement a 3D cube masking strategy that enables efficient processing of tensor-like spectral data. Our approach leverages a 90\% masking rate to capture spatially and spectrally visual representations in an effective way, leading to more accurate and diverse knowledge extraction from input spectral data.

Given a 3D cube-shaped spectral image $\bm{x}\in \mathbb{R}^{H\times W\times D}$, we partition it into non-overlapping 3D tensor tokens along both the spatial and spectral dimensions. Each token has a size of $p\times p\times k$, where $p$ and $k$ are the token sizes in spatial and spectral dimensions, respectively. Using these settings, we then have $\frac{H}{p}\times \frac{W}{p}\times \frac{D}{k}$ tokens, denoted as $\bm{x}=\{\bm{x}_{1}, ..., \bm{x}_{i},...,\bm{x}_{\frac{H}{p}\times \frac{W}{p}\times \frac{D}{k}}\}$. This results in the following visible and masked representations (e.g., $\bm{x}_{vis}$ and $\bm{x}_{mask}$) on the spectral data.
\begin{equation}
\label{eq1}
\begin{aligned}
       [\bm{x}_{vis},\bm{x}_{mask}]=\mathbb{M}\odot\bm{x},
\end{aligned}
\end{equation} 
where $\mathbb{M}\in \{0,1\}^{\frac{H}{p}\times \frac{W}{p}\times \frac{D}{k}}$ is a token-wise binary mask indicating which tokens should be masked, i.e., all pixels in the token are set to zero. Accordingly, $\odot$ denotes the masking operation.

\subsection{Encoder for Visible Tokens}
Similar to the encoder in MAE, all visible tokens $\{\bm{x}_{i}|i\in vis\}$ spatial-spectral mixed representations are first transformed into feature embeddings using a shared linear projection $\bm{E}_{s}$. The learned representations via the encoder $f_{\theta}$ with respect to the variable $\bm{\theta}$ are denoted as $f_{\bm{\theta}}(\bm{E}_{s}\bm{x}_{i}+\bm{E}_{pos})$, where $\bm{E}_{pos}$ represents the positional encoding. The encoder $f_{\theta}$ comprises several stacking self-attention (SA) transformer blocks. The SA module used in the encoder can be constructed as follows. 

\begin{itemize}
    \item The input embeddings $\bm{z}_{i}$ are linearly transformed to {\it query} $\bm{Q}_{i}$,  {\it key} $\bm{K}_{i}$, and  {\it value} $\bm{V}_{i}$ embeddings using learnable projection matrices $\bm{W}_{Q}$, $\bm{W}_{K}$, and $\bm{W}_{V}$, respectively.
    \item The attention scores $\bm{S}_{i}$ between {\it query} and  {\it key} embeddings are computed as a dot product scaled by $\frac{1}{\sqrt{d}}$ and passed through a softmax function. The resulting scores are then used to weight the  {\it value} embeddings, which are summed to produce the final output embeddings, i.e., $\bm{z}_{i}=\bm{S}_{i}\bm{V}_{i}$. The formula of the original SA's complete process can be written as
    \begin{equation}
    \label{eq2}
    \begin{aligned}
         &\bm{Q}_{i}=\bm{x}_{i}\bm{W}_{Q},\; \bm{K}_{i}=\bm{x}_{i}\bm{W}_{K},\;
         \bm{V}_{i}=\bm{x}_{i}\bm{W}_{V},\\
         &\bm{S}_{i}={\rm softmax}(\frac{\bm{Q}_{i}\bm{K}_{i}^{\top}}{\sqrt{d}}),\\
         &\bm{z}_{i}={\rm Attention}(\bm{Q}_{i},\bm{K}_{i}, \bm{V}_{i})=\bm{S}_{i}\bm{V}_{i},
    \end{aligned}
    \end{equation}
    where $d$ denotes the dimension of embeddings.
    \item Finally, the output features $\bm{z}_{i}$ have the same dimension as $\bm{x}_{i}$ and can be further processed by subsequent encoders.
\end{itemize}

\subsection{Lightweight Decoder with Multi-Target Reconstruction}
Given the encoder output features $\bm{z}$, we simultaneously train a lightweight decoder with a multi-target reconstruction strategy with respect to the variable $\bm{\phi}$ on $\bm{z}$ to recover the original image tokens from potential embeddings of visible and masked image tokens. Mathematically, the reconstructed image tokens $\bm{\hat{x}}$ can be formulated as $\bm{\hat{x}}=g_{\bm{\phi}}(f_{\bm{\theta}}(\mathbb{M}\odot\bm{x}))$. The decoder $g_{\bm{\phi}}$ is typically narrower and shallower than the encoder, and usually consists of a few transformer blocks and a linear reconstruction layer. The proposed SpectralGPT trains the encoder $f_{\bm{\theta}}$ and decoder $g_{\bm{\phi}}$ in an end-to-end manner to minimize the reconstruction loss between the reconstructed image tokens $\bm{\hat{x}}$ and the original image tokens $\bm{x}$. 
In our approach, the reconstruction loss consists of two components: token-to-token and spectral-to-spectral. This multi-target reconstruction allows the learned representations to capture spatial-spectral coupling characteristics and spectrally sequential information effectively. This overall loss $\mathcal{L}$ is quantified mathematically using MSE in the pixel space, defined as follows:
    \begin{equation}
    \label{eq3}
    \begin{aligned}
        \mathcal{L}&=\mathcal{L}_{token}+\lambda\mathcal{L}_{spectral}\\
          &=\frac{1}{m}\sum_{i\in vis}(\bm{x}_{i}-\bm{\hat{x}}_{i})^{2}+\frac{1}{n}\sum_{j=1}^{n}(\bm{x}_{j}-\bm{\hat{x}}_{j})^{2}\\
          &=\frac{1}{m}\sum_{i\in vis}(\bm{x}_{i}-\bm{\hat{x}}_{i})^{2}\\
          &\;\;+\frac{1}{n}\sum_{j=1}^{n}([\bm{x}_{r,c,1}, ...,\bm{x}_{r,c,\frac{D}{k}}]_{j}-[\bm{\hat{x}}_{r,c,1},...,\bm{\hat{x}}_{r,c,\frac{D}{k}}]_{j})^{2},\\
    \end{aligned}
    \end{equation}
where $m$ and $n$ represent the number of masked tokens ($\frac{H}{p}\times \frac{W}{p}$) and spectral tokens (each spectral token consists of $\frac{D}{k}$ standard tokens along with spectral dimension), and $(r,c)$ denotes the standard token in the $i$-th row and the $j$-th column of the spectral data. 

\subsection{Progressive Pretraining}
The proposed SpectralGPT model has the advantage of being highly adaptable to different input image sizes, which is especially useful for processing large datasets with images of varying size, resolution, temporal variability, and geographical coverage. This is achieved by dividing the input image into fixed-sized 3D tokens (e.g., $8\times 8\times 3$), which are then independently processed through the encoder-decoder pipeline. The resulting tokens are then stitched back together to form the final output image. This approach ensures that the model can handle images of arbitrary dimensions in theory, without requiring any changes to the architecture or hyperparameters. With this characteristic, the proposed model allows for feeding varying-sized images into the encoder networks and also might enable the input images with different sensors, resolutions, time series, and modalities, as long as the 3D tokens are cropped to a fixed size.

It is worth emphasizing that the progressive feeding of different types of input images into the networks is not only useful for enabling greater flexibility in the type and size of input images but also for improving the model's ability to extract valuable knowledge from diverse data sources, thereby enhancing model generalization. This can be achieved, for instance, by first inputting images with a size of $96\times 96$ pixels and then progressively feeding in the images with the size of $128\times 128$ pixes, or by starting with Sentinel-2 data and then transitioning to Landsat-8 or Gaofen-2 data. More broadly, the ability to process varying types and sizes of input images can lead to more robust and generalizable features that are not limited to a specific input image type or size, thus improving model generalization and performance on previously unseen data. Moreover, this flexibility in input image size and type is particularly beneficial in practical applications, where input images may come from different sources or have varying resolutions, allowing the model to be more adaptable to real-world scenarios where input images can be unpredictable.

The input order of the datasets (i.e., feeding fMoW-S2 prior to BigEarthNet-S2) is arranged in the progressive training strategy. The motivation for such design is motivated by two primary considerations. On one hand, the quality of spectral data in fMoW-S2 is relatively lower than that in BigEarthNet-S2, whereas its scale is much larger. As a result, the model benefits from initiating learning from data with comparatively lower quality but large scale, followed by fine-tuning with higher-quality data. On the other hand, the image size of fMoW-S2 is smaller than that of BigEarthNet-S2. We found that the model tends to initially learn from the small-size data and further extend to the larger one. Furthermore, the order setting not only expedites network convergence but also facilitates higher-efficiency training in networks toward a better solution.

\begin{figure}[!t]
      \centering
	   \includegraphics[width=0.5\textwidth]{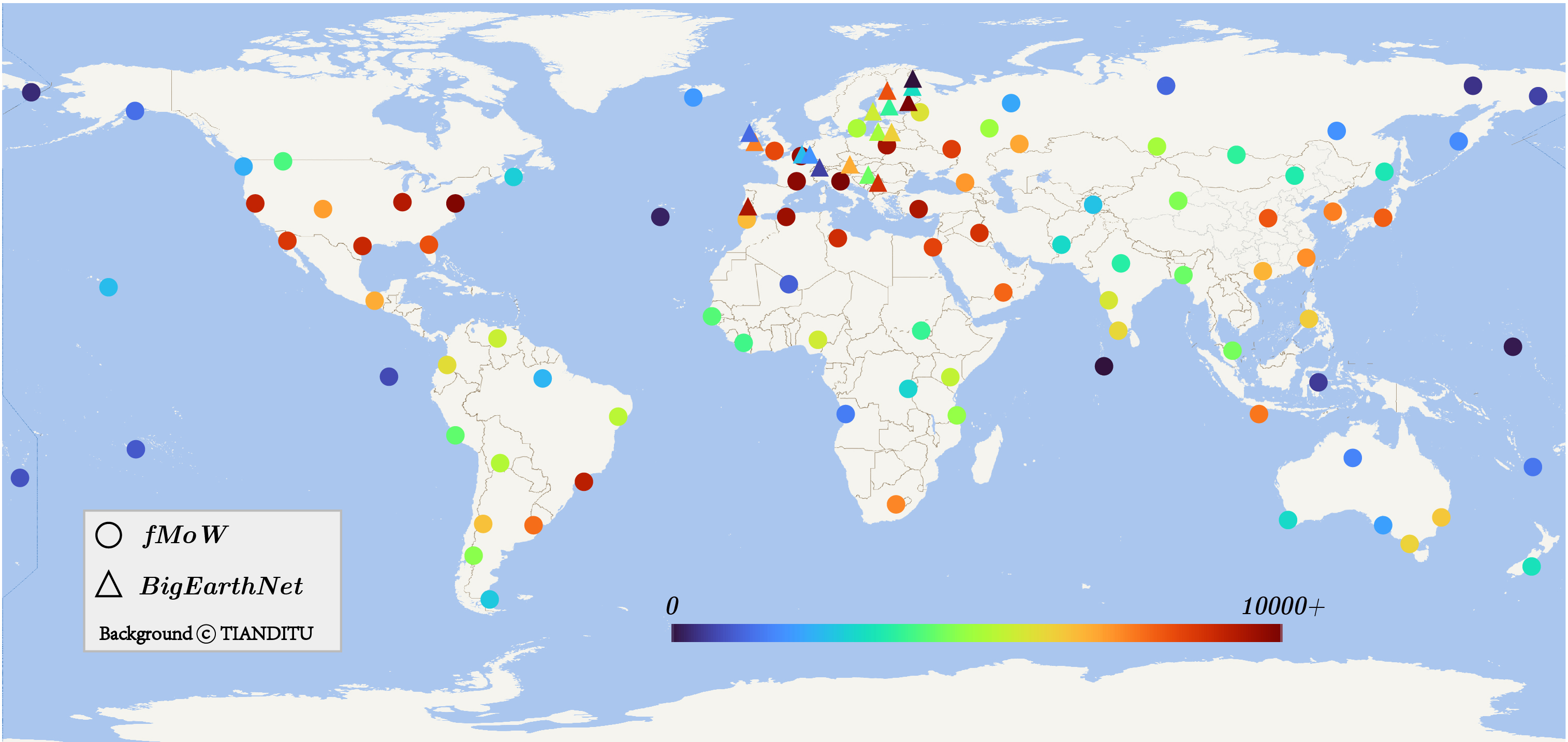}
      \caption{Sample density distribution collected from Sentinel-2 sources (i.e., fMoW, BigEarthNet) over the Earth's inhabited areas, amounting to a total of 1,473,105 images.}
\label{fig:data_overview}
\end{figure}

\subsection{Pretrained Dataset}
Our foundation model is trained on a comprehensive dataset comprising over one million spectral images from the Sentinel-2 satellite. This dataset encompasses 12 spectral bands and draws from two primary sources: fMoW-S2 \cite{cong2022satmae}, a globally diverse collection labeled with 62 categories based on the Functional Map of the World (fMoW) \cite{christie2018functional}, and BigEarthNet \cite{sumbul2019bigearthnet}, a regional dataset originating from over ten European countries. To provide an overview of the dataset, Fig. \ref{fig:data_overview} illustrates the distribution of image samples over Earth's inhabited areas, amounting to a total of 1,473,105 images.

Researchers at Stanford University meticulously curated a dataset by leveraging geo-coordinates and timestamps from the fMoW dataset. This process aims to construct a time series of Sentinel-2 images. To ensure data quality, locations with exclusively pre-Sentinel-2 fMoW images were excluded. For locations with partially preceding fMoW images, selective curation is performed, involving the exclusion of these specific images and the introduction of supplementary captures at 6-month intervals to enrich the temporal sequence. This approach culminates in the creation of the fMoW Sentinel-2 dataset, denoted as fMoW-S2. This dataset predominantly covers fMoW locations and preserves labels mirroring those in the original fMoW dataset.

The fMoW-S2 dataset comprises Sentinel-2 spectral images (B1-12 and B8A) and is partitioned into three subsets: 712,874 training images, 84,939 validation images, and 84,966 test images, totaling 882,779 images. Each image has an average dimension of approximately 45 pixels in height and 60 pixels in width. For additional details about the fMoW-S2 dataset, interested parties can refer to the dedicated website\footnote{https://purl.stanford.edu/vg497cb6002}.

Furthermore, the study incorporates the BigEarthNet dataset, specifically the BigEarthNet-S2 variant\footnote{https://bigearth.net}, comprising 590,326 distinct, non-overlapping Sentinel-2 spectral image tokens. The pretraining of different variant models using the proposed method involves the use of 712,874 fMoW-S2 images and 354,196 BigEarthNet-S2 images. Notably, only 10\% of the BigEarthNet-S2 images with labels, amounting to 35,420 images, are utilized for fine-tuning in downstream tasks.

\subsection{Implementation Details and Experimental Setup}\label{pretrain_imple}
Following established conventions, we acknowledge that Sentinel-2 images comprise 13 spectral bands. However, to harmonize datasets across pretrained and downstream tasks in terms of channel composition, we have opted to retain the 12 dominant bands, excluding band B10, in all fMoW dataset images. To ensure data consistency, we normalize the spectral images band by band, scaling their values to a standardized range of 0 to 1. Subsequently, we follow established methodologies \cite{cong2022satmae} for preprocessing. This involves random image cropping within the range of 0.2x to 1.0x of the original size, resizing them to $96\times96$ pixels, and applying horizontal flips. These meticulous steps collectively contribute to the robustness and compatibility of our spectral foundation model.

We employ the vanilla ViT-Base architecture as the network backbone. To adapt the model to spectral data, we employ a token size of $8\times 8\times 3$ pixels, effectively partitioning the images. For instance, an image with a size of $96\times 96\times 12$ pixels is segmented into $12\times 12\times 4$ tokens. Drawing inspiration from a prior work \cite{feichtenhofer2022masked}, our approach incorporates two learnable positional embeddings. One of these embeddings is dedicated to spatial information, while the other is tailored to capture variations across spectral channels. This augmentation further refines the model's ability to extract meaningful features from the spectral input.

Our pretraining closely adheres to the approach outlined in a prior study \cite{cong2022satmae}. Utilizing the computational power of 8 NVIDIA GeForce RTX 4090 GPUs and AMD EPYC 7Y83 CPU, we implement the AdamW optimizer \cite{loshchilov2019decoupled} with a foundational learning rate of $10^{-4}$, coupled with a half-cycle cosine decay schedule. To ensure robustness, we adopt a 3D masking ratio of 90\%, facilitating effective training. The model undergoes a comprehensive pretraining regimen spanning 200 epochs on the fMoW-S2 dataset. After this phase, the model's training continues on the BigEarthNet-S2 dataset for 100 epochs. While this phase necessitates a modification in input dimensions to $128\times 128 \times 12$, the other settings remain consistent. This meticulous strategy effectively enhances the model's adaptability and performance across diverse datasets. 

\section{Experiments}  
\label{experiments}
In this section, we rigorously evaluate the performance of our SpectralGPT model by benchmarking it against several SOTA foundation models: ResNet50 \cite{he2016deep}, SeCo \cite{manas2021seasonal}, ViT \cite{dosovitskiy2020image}, and SatMAE \cite{cong2022satmae}. Further, we assess its capabilities across four downstream EO tasks, including single-label scene classification, multi-label scene classification, semantic segmentation, and change detection, as well as extensive ablation studies. 

We quantitatively assess the performance of pretrained foundation models across four downstream tasks in terms of recognition accuracy for the single-label RS scene classification task, macro and micro mean average precision (mAP), i.e., macro-mAP (micro-mAP), for the multi-label RS scene classification task, overall accuracy (OA) and mean intersection over union (mIoU) for the semantic segmentation task, and precision, recall, and F1 score for the change detection. Additionally, we conduct insightful ablation studies, exploring critical factors such as masking ratio, decoder depth, model size, patch size, and training epochs. Utilizing the computational power of 4 NVIDIA GeForce RTX 4090 GPUs, we meticulously fine-tune pretrained foundation models for both downstream tasks and ablation studies, thereby offering comprehensive insights into SpectralGPT's capabilities and adaptability within the RS domain.

\begin{figure}[!t]
      \centering	   
      \includegraphics[width=0.485\textwidth]{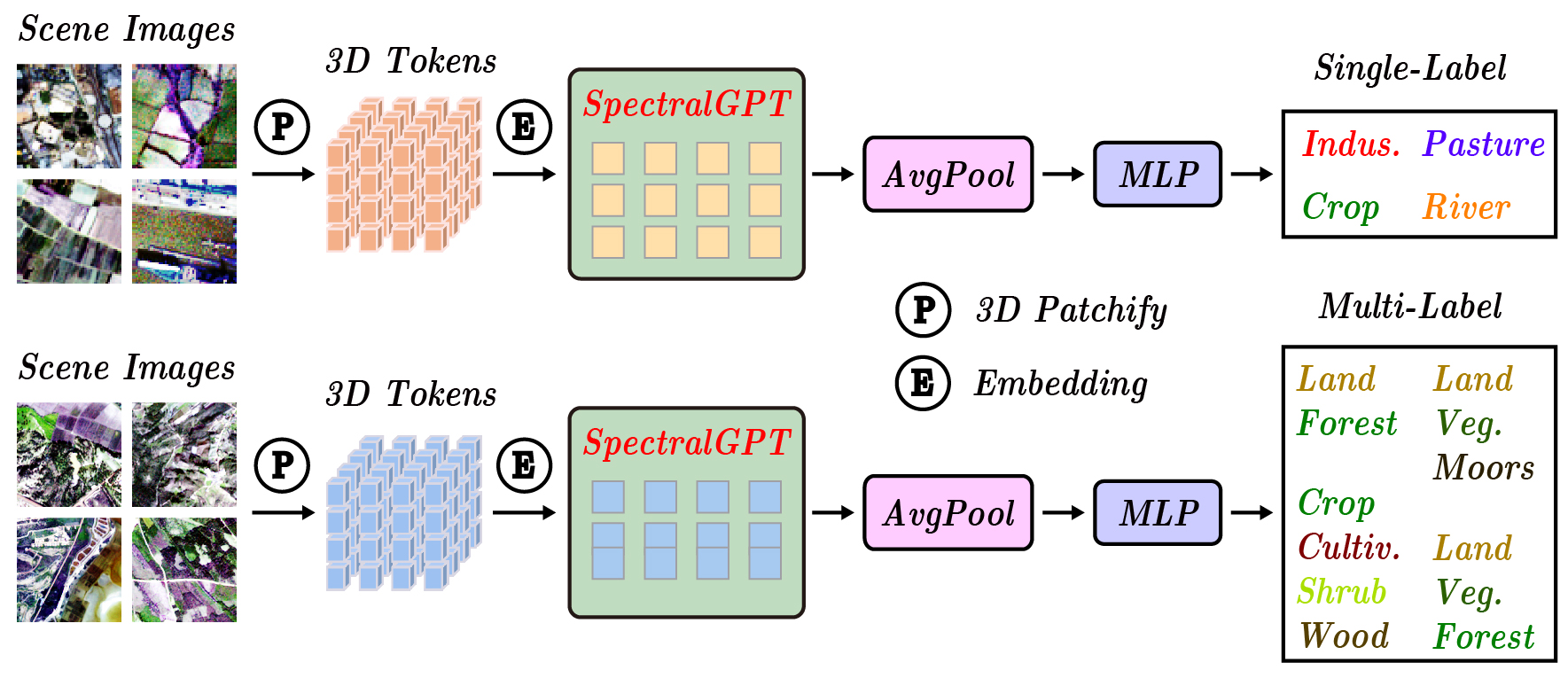}
      \caption{Network architecture for downstream tasks in terms of single-label (Top) RS scene classification and multi-label (Bottom) RS scene classification by leveraging our pretrained SpectralGPT model. The \textit{AvgPool} and \textit{MLP} denote the average pooling operation and the multilayer perception, respectively.}
\label{fig:Downstream-CL}
\end{figure}

\subsection{Single-Label RS Scene Classification on EuroSAT}

For the downstream single-label RS scene classification task, we employ the EuroSAT dataset \cite{helber2019eurosat}. This dataset consists of 27,000 Sentinel-2 satellite images collected from 34 European countries. These images are classified into 10 land use classes, each containing between 2,000 to 3,000 labeled images. Each image in this dataset has a resolution of $64\times 64$ pixels and encompasses 13 spectral bands. It's worth noting that, for consistency with prior data processing, band B10 has been excluded from all images. Additionally, we follow the train/validation splits as recommended in \cite{neumann2020domain}.

On the EuroSAT dataset, these pretrained models undergo fine-tuning, spanning 150 epochs with a batch size of 512. This fine-tuning process employs the AdamW optimizer with a base learning rate of $2\times10^{-4}$, and it incorporates data augmentations in line with prior work \cite{he2022masked}, including weight decay (0.05), drop path (0.1), reprob (0.25), mixup (0.8), and cutmix (1.0). The foundational encoder of the pretrained model is utilized, and its output is passed through an average pooling layer to generate predictions. The training objective is to minimize the cross-entropy loss. Fig. \ref{fig:Downstream-CL} illustrates the network architecture for the downstream single-label scene classification task.
  
The pretrained model's encoder serves as the foundational backbone, and its output is subject to an average pooling layer to generate predictions. The training objective involves minimizing the cross-entropy loss. In Table \ref{eurosat}, we present a comparative analysis of our proposed method against alternative pretraining models, reporting the highest Top1 accuracy on the validation set. The obtained results highlight the efficacy of the proposed approach, achieving an impressive accuracy of 99.15\%. Furthermore, when the model undergoes pretraining on both the fMoW-S2 and BigEarthNet datasets, a noteworthy performance boost is observed, culminating in a remarkable accuracy of 99.21\%. This underscores the advantage of leveraging diverse data sources for improved model performance.

\begin{table}[!t]
    \begin{center}
	\caption{Quantitative results of SOTA pretrained foundation models for the downstream single-label RS scene classification task in terms of recognition accuracy on the EuroSAT dataset. The best result is shown in bold.}
        \label{eurosat}
	\begin{tabular}{ccc}
	\toprule[1.5pt]
	{Method} & {Pretrained Dataset} & {Acc. (\%)} \\
        \hline\hline        
        ResNet50 \cite{he2016deep} & {ImageNet-1k} & {96.72} \\
        SeCo \cite{manas2021seasonal} & SeCo & 97.23\\
        ViT \cite{dosovitskiy2020image} & Random Init. & 98.73 \\ 
        ViT-22k \cite{dosovitskiy2020image} & ImageNet-22k & 98.91 \\
        SatMAE \cite{cong2022satmae} & fMoW-S2 & 99.09 \\
        \hline
        SpectralGPT & fMoW-S2 & 99.15 \\
        SpectralGPT$^{+}$ & fMoW-S2+BigEarthNet & \textbf{99.21} \\
	\bottomrule[1.5pt]
	\end{tabular}
	\label{tab:synthetic}
	\end{center}
\end{table}

\subsection{Multi-Label RS Scene Classification on BigEarthNet} \label{subsection_bigearthnet}
For the multi-label RS scene classification task, we utilize the BigEarthNet-S2 dataset \cite{sumbul2019bigearthnet}. This extensive dataset consists of 125 Sentinel-2 tiles and comprises 590,326 12-band images that span 19 classes for multi-label classification. The images encompass resolutions ranging from 10 to 60 meters, with 12\% of low-quality images being excluded. The training and validation sets align with prior research \cite{neumann2020domain}, with 354,196 training samples and 118,065 validation samples. To prepare the images for model training, those of varying resolutions are standardized to a uniform size of $128\times 128$ pixels using bilinear interpolation.

On the BigEarthNet-S2 dataset, these foundation models are fine-tuned using a 10\% subset of the training data, following similar settings to those applied in the EuroSAT fine-tuning experiments, except for an increased learning rate of $2\times10^{-4}$, which is in line with findings from previous research \cite{manas2021seasonal,cong2022satmae}. Most existing methods, including those that use pretrained foundation models, typically utilize all available images for training in the BigEarthNet-S2 dataset. In contrast, our proposed SpectralGPT achieves higher classification performance, even when utilizing only 10\% of the training samples. Given the multi-label classification nature of this task, our training objective involves the multi-label soft margin loss, and performance evaluation is based on the mAP metric. Notably, we calculate mAP using both macro and micro mAP measurements. This approach is particularly relevant for the BigEarthNet-S2 dataset, which exhibits class imbalance. The multi-label classification framework is shown in Fig. \ref{fig:Downstream-CL}. 

Table \ref{bigearthnet} presents a comparative analysis of our pretrained model against other proposed pretrained models and models trained from scratch, showcasing the exceptional performance of the proposed approach. In particular, when compared to ViT pretrained on ImageNet-22k and SatMAE, our SpectralGPT model outperforms them by 0.84\% (0.82\%) and 0.71\% (0.68\%) in terms of macro-mAP (micro-mAP), respectively. Notably, the introduction of additional pretraining data (BigEarthNet), i.e., SpectralGPT$^+$, leads to a significant performance boost, with the model achieving an impressive 88.22\% (87.50\%) macro-mAP (micro-mAP), surpassing the model solely trained on fMoW-S2 by 2.19\% (1.86\%). This substantial improvement can be attributed to two key factors. Firstly, the model's initial pretraining on BigEarthNet (even without labels) equips it with a strong grasp of the dataset's distribution, accelerating convergence during fine-tuning and enhancing mAP. Secondly, the adoption of the MIM method as a pretraining pretext task, coupled with a substantial data scale, necessitates alignment with the training strategy, emphasizing the significance of the random masking framework and a 90\% masking ratio to facilitate more robust representation learning. Furthermore, as our evaluation focuses on a multi-label classification task and employs only 10\% of the training data, the results underscore the superior generalization and few-shot learning capabilities of our proposed model in tackling challenging downstream tasks.

\begin{table}[!t]
    \begin{center}
	\caption{Quantitative results of SOTA pretrained foundation models for the downstream multi-label RS scene classification task in terms of mean average precision (mAP) on the BigEarthNet dataset. The best result is shown in bold.}
    \resizebox{0.5\textwidth}{!}{
	\begin{tabular}{cccc}
	\toprule[1.5pt]
	Method & Pretrained Dataset & macro-mAP & micro-mAP\\
        \hline\hline
        ResNet50 \cite{he2016deep} & ImageNet-1k & 80.76 & 80.06 \\
        SeCo \cite{manas2021seasonal} & SeCo & 83.16 & 82.82 \\
        ViT \cite{dosovitskiy2020image}& Random Init. & 81.57 & 80.15 \\
        ViT-22k \cite{dosovitskiy2020image} & ImageNet-22k & 85.08 & 84.67 \\
        SatMAE \cite{cong2022satmae} & fMoW-S2 & 85.21&84.93\\
        \hline
        SpectralGPT & fMoW-S2 & 86.03&85.61 \\
        SpectralGPT$^{+}$ & fMoW-S2+BigEarthNet & \textbf{88.22} &\textbf{87.50} \\
	\bottomrule[1.5pt]
	\end{tabular}}
	\label{bigearthnet}
	\end{center}
\end{table}

\begin{figure}[!t]
      \centering	   
      \includegraphics[width=0.485\textwidth]{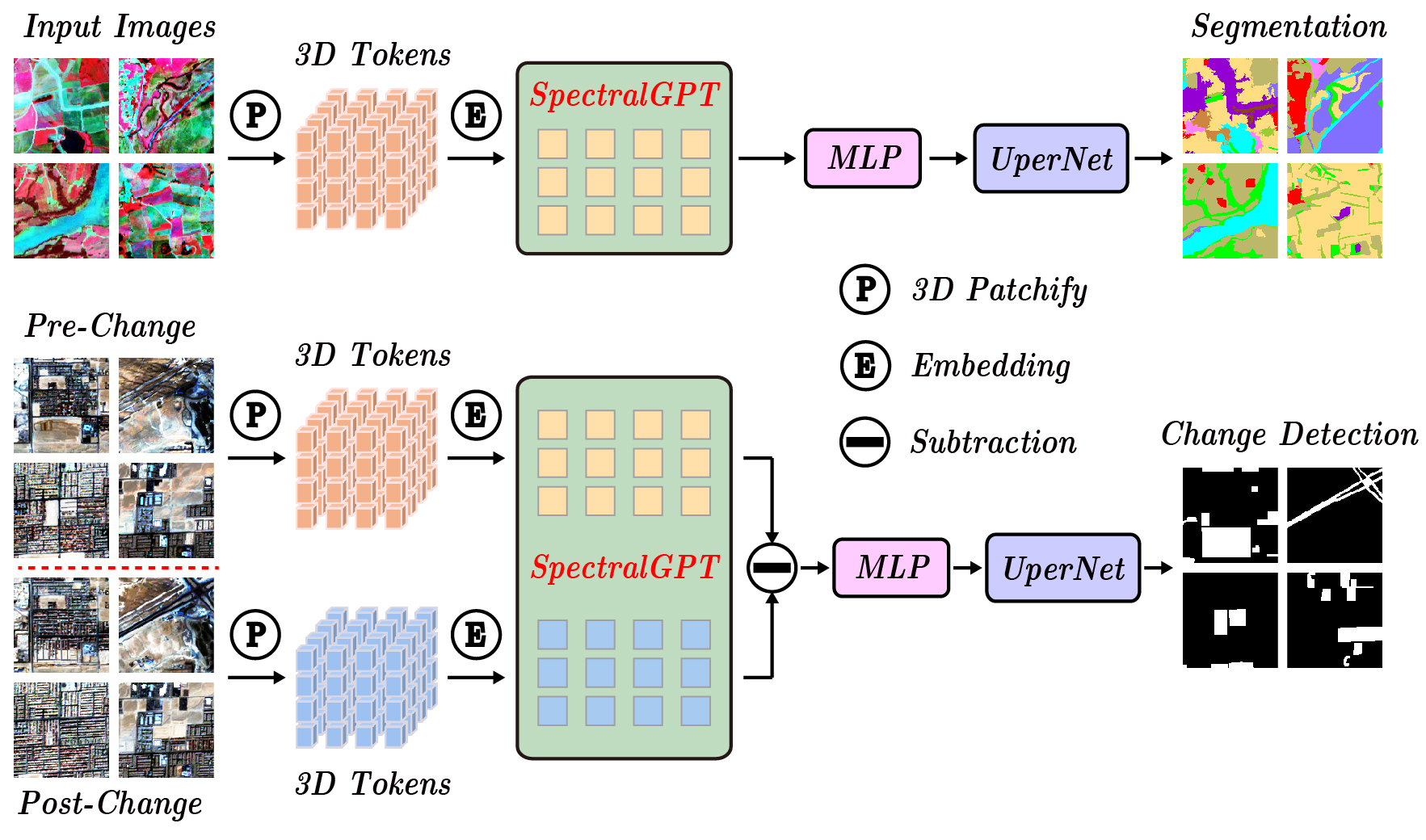}
      \caption{Network architecture for downstream tasks in terms of semantic segmentation (Top) and change detection (Bottom) by leveraging our pretrained SpectralGPT model and training a follow-up UperNet \cite{xiao2018unified} Head. The \textit{MLP} denotes the multilayer perception.}
\label{fig:Downstream-CD-Seg}
\end{figure}

\begin{figure*}[!t]
      \centering
      \includegraphics[width=1\textwidth]{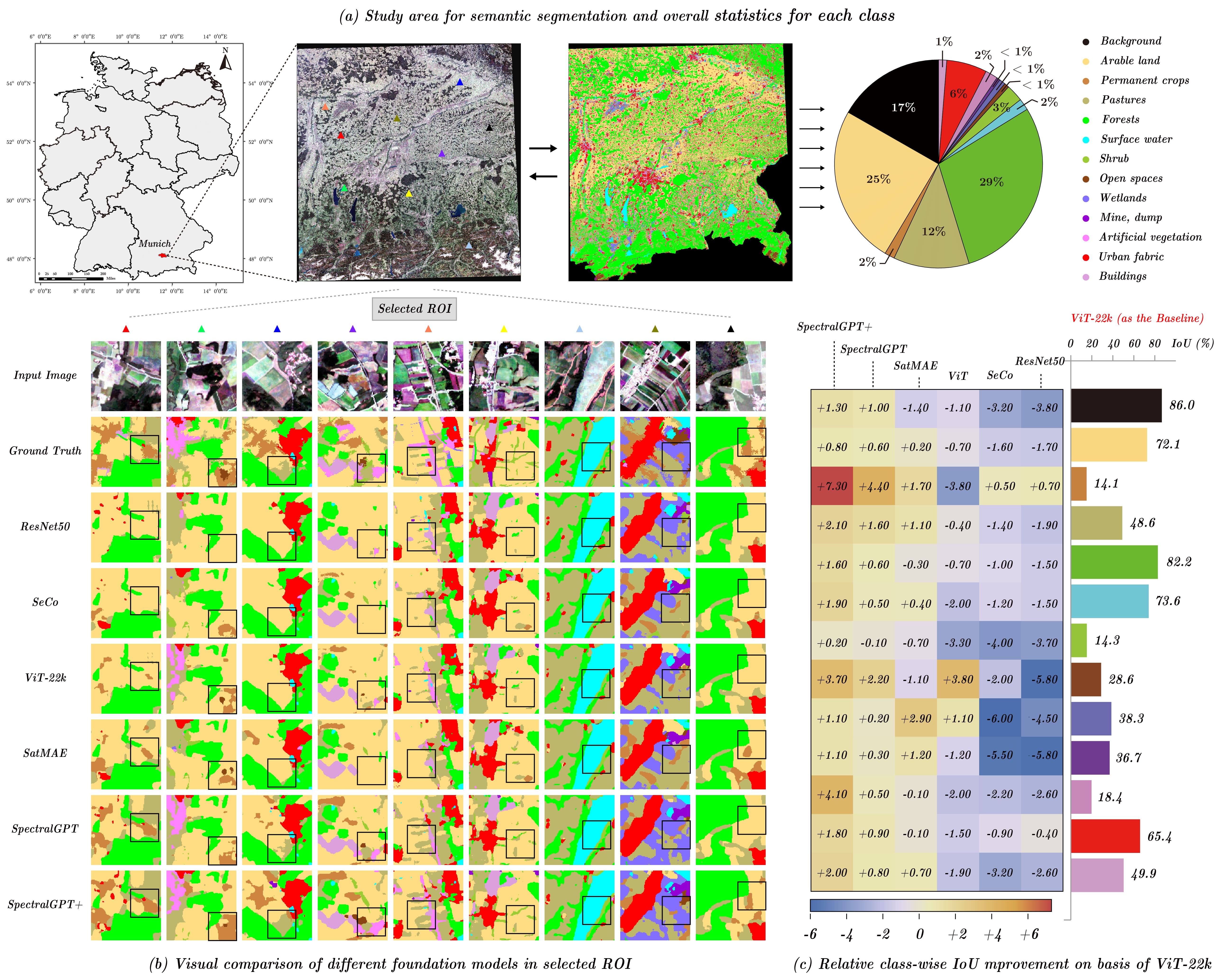}
      \caption{Qualitative and quantitative semantic segmentation results of different pretrained foundation models on the SegMunich dataset. (a) Study area and overall statistics for each class in the semantic segmentation task. (b) Segmentation visualization maps of different foundation models in selected ROIs. (c) Relative IoU performance improvement for each class on the basis of ViT-22k.}
\label{fig:SegMunich}
\end{figure*}

\subsection{RS Semantic Segmentation on SegMunich} \label{subsection_seg}

For the semantic segmentation task, we create a new SegMunich dataset, which is derived from the Sentinel-2 spectral satellite \cite{hong_danfeng_2023_8412377}. This dataset consists of a 10-band best-pixel composite with dimensions of $3,847\times 2,958$ pixels and a spatial resolution of 10 meters. It captures Munich's urban landscape over a span of three years up to April 2020 and includes a segmentation mask that meticulously delineates 13 Land Use and Land Cover (LULC) classes. The data for this mask is sourced from various places, including OpenStreetMap for street network data and the OSMLULC platform\footnote{https://osmlanduse.org/} for the remaining 12 classes, all obtained at the same 10-meter spatial resolution. To create a comprehensive feature representation for semantic segmentation, the dataset combines the 10-meter spectral bands (B1, B2, B3, and B4) with resampled 20-meter spectral bands (B5, B6, B7, B8A, B11, B12), which have been upsampled to match the 10-meter resolution. This amalgamation of spectral bands ensures that the dataset provides rich and informative data for the semantic segmentation task.

On the SegMunich dataset, we employ the UperNet framework \cite{xiao2018unified} in conjunction with the pretrained foundation models, initially consolidating the four tokens per pixel from the encoder's final layer into a single token. Image data is divided into $128\times 128$-pixel tokens with a 50\% overlap. The dataset is then split into a training-validation ratio of 8:2 and subjected to data augmentation techniques, including random flips and rotations. During fine-tuning on this dataset, we use a batch size of 96 and set the base learning rate to $5\times10^{-4}$. The optimization and loss functions remain consistent with those employed in the EuroSAT experiment, ensuring a coherent and uniform approach to model training and evaluation. The segmentation architecture is detailed in Fig. \ref{fig:Downstream-CD-Seg}. 

Table \ref{segmentation} lists quantitative results in terms of OA and mIoU for the semantic segmentation task. Our SpectralGPT (SpectralGPT$^+$) outperforms all others, exhibiting a significant lead with a 1.1\% (2.3\%) higher mIoU than the second-best result (i.e., SatMAE). Fig. \ref{fig:SegMunich}(a) offers a visual depiction of the Munich area under study for the segmentation task, along with the proportions of the 13 classes. The qualitative comparison in several ROIs, as shown in Fig. \ref{fig:SegMunich}(b), highlights our model's superior ability to recognize a wider range of land use categories well compared to the competing models in most instances. Furthermore, when considering ViT-22k as the baseline for performance comparison, our model consistently excels across all segmentation classes, as evident in Fig. \ref{fig:SegMunich}(c), particularly for categories, such as \textit{crops}, \textit{pastures}, \textit{open spaces}, \textit{vegetations}, and others. By amalgamating category statistics with class-wise IoU outcomes, it becomes evident that our SpectralGPT model excels in mitigating the challenges posed by category-imbalanced classification. This results in a substantial enhancement in performance when compared to other foundation models.

\begin{table*}[!t]
       \centering
	\caption{Quantitative results of SOTA pretrained foundation models that are finely tuned using UperNet for the downstream RS semantic segmentation task in terms of overall accuracy (OA) and mean intersection over union (mIoU) on the SegMunich dataset. The best result is shown in bold.}
     \resizebox{1\textwidth}{!}{
	\begin{tabular}{c|c|c|ccccccccccccc|c}
	   \toprule[1.5pt]
		 {Method} &{Pretrained Dataset} &{OA}&\rotatebox{75}{Background}&\rotatebox{75}{Arable land}&\rotatebox{75}{Perm. Crops}&\rotatebox{75}{Pastures}&\rotatebox{75}{Forests}&\rotatebox{75}{Surface water}&\rotatebox{75}{Shrub}&\rotatebox{75}{Open spaces}&\rotatebox{75}{Wetlands}&\rotatebox{75}{Mine, dump}&\rotatebox{75}{Artificial veg.}&\rotatebox{75}{Urban fabric}&\rotatebox{75}{Buildings}&{mIoU} \\
        \hline\hline       
        ResNet50 \cite{he2016deep} & ImageNet-1k &80.1&82.2&70.4&14.8&46.7&80.7&72.1&10.6&22.8&33.8&30.9&15.8&65.0&47.3&45.6 \\
        SeCo \cite{manas2021seasonal} & SeCo &80.3&82.8&70.5&14.6&47.2&81.2&72.4&10.3&26.6&32.3&31.2&16.2&64.5&46.7&45.9 \\
        ViT \cite{dosovitskiy2020image} & Random Init &81.0&84.9&71.4&10.3&48.2&81.5&71.6&11.0&\textbf{32.4}&39.4&35.5&16.4&63.9&48.0&47.3 \\
        ViT-22k \cite{dosovitskiy2020image} & ImageNet-22k &81.7&86.0&72.1&14.1&48.6&82.2&73.6&14.3&28.6&38.3&36.7&18.4&65.4&49.9&48.3 \\
        SatMAE \cite{cong2022satmae} & fMoW-S2 &81.5&84.6&72.3&15.8&49.7&81.9&74.0&13.6&27.5&\textbf{41.2}&\textbf{37.9}&18.3&65.3&50.6&48.7\\
        \hline
        SpectralGPT & {fMoW-S2} &82.5&87.6&73.1&16.3&50.6&83.6&74.7&14.2&32.5&39.2&37.7&19.4&67.0&51.7&49.8 \\
        SpectralGPT$^{+}$ & fMoW-S2+BigEarthNet &\textbf{82.7}&\textbf{88.0}&\textbf{73.1}&\textbf{22.5}&\textbf{51.0}&\textbf{84.1}&\textbf{76.0}&\textbf{14.5}&\textbf{33.7}&39.6&\textbf{38.7}&\textbf{22.5}&\textbf{67.4}&\textbf{52.0}&\textbf{51.0} \\
	\bottomrule[1.5pt]
	\end{tabular}}
	\label{segmentation}
\end{table*}

\begin{figure*}[!t]
      \centering	   
      \includegraphics[width=1\textwidth]{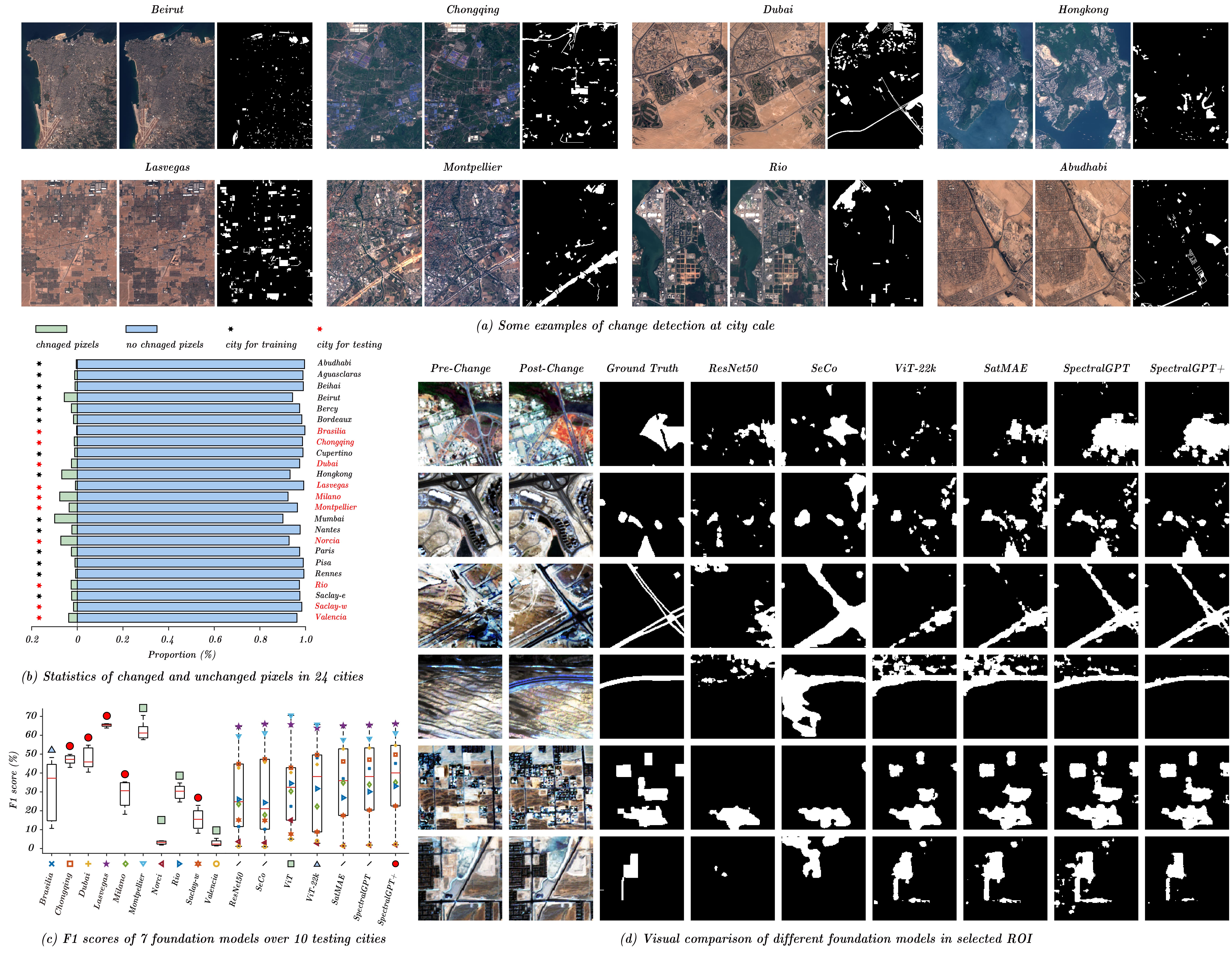}
      \caption{Qualitative and quantitative change detection results of different pretrained foundation models on the OSCD dataset. (a) Some studied examples of change detection at the city scale. (b) Statistics of changed and unchanged pixels in 24 studied cities, including training and testing scenes. (c) F1 scores of different foundation models over 10 testing cities. (d) Change detection visualization maps of different foundation models in selected ROIs.}
\label{fig:CD}
\end{figure*}

\subsection{RS Change Detection on OSCD}
For the change detection task, we utilize the OSCD dataset \cite{daudt2018urban}. Fig. \ref{fig:CD}(a) shows several examples at the city scale. This dataset comprises 24 cities of Sentinel-2 images, with 14 images used for training and 10 images for evaluation. These images were captured between 2015 and 2018 and encompassed 13 spectral bands with resolutions of 10m, 20m, and 60m. The dataset is annotated at the pixel level to indicate changes, specifically focusing on urban developments.

On the OSCD dataset, we perform image cropping to create patches of size $128\times 128$ pixels with a 50\% overlap rate, and we apply random flips and rotations as data augmentation techniques. For each pair of images, both are simultaneously processed through a shared encoder, and the difference between their features is computed and then passed to a UperNet. Each feature pixel consists of 4 tokens, similar to the segmentation approach, and we use a linear layer to consolidate these 4 tokens into 1 token. The model is trained for 60 epochs with a batch size of 64 and a learning rate set to $1\times10^{-3}$, using negative log-likelihood loss as the training objective. The entire framework for leveraging the pretrained SpectralGPT model in the change detection task is depicted in Fig. \ref{fig:Downstream-CD-Seg}.

Model performance is assessed through precision, recall, and F1 score, with quantitative outcomes shown in Table \ref{changedetection} on the OSCD dataset, where our proposed model achieves the highest F1 score, surpassing the second-best model (i.e., SatMAE) by a substantial margin of 0.75\% (1.53\%). However, it is worth noting that our model excels in F1 score and recall but exhibits relatively lower precision compared to other models. This phenomenon can be attributed to two main factors. Firstly, extreme imbalance of the inherent data within the change detection task (see Fig.\ref{fig:CD}(b)), where the number of positive and negative samples varies significantly, may lead the model to classify negative cases as positive to improve recall at the cost of precision. Secondly, the complexity of the ViT architecture demands a substantial amount of data to mitigate overfitting. The model may struggle with overfitting and become less adaptable to out-of-domain data. Addressing this challenge could involve providing additional fine-tuning data or reducing the model's rank. In terms of qualitative results, our model excels in predicting change pixels with fewer false negatives in selected ROIs of Fig. \ref{fig:CD}(d). Significantly, Fig. \ref{fig:CD}(c) accentuates the exceptional performance of SpectralGPT, with our model achieving the top results in half of the testing cities. In addition, there is a consistent performance trend among the compared foundation models across 10 different testing cities, with \textit{Lasvegas} and \textit{Montpellier} consistently achieving the highest and second-highest F1 scores, respectively.

\begin{table}[!t]
   \centering
   \caption{Quantitative results of SOTA pretrained foundation models that are finely tuned using UperNet for the downstream RS change detection task in terms of precision, recall, and F1 score on the OSCD dataset. The best result is shown in bold.}
    \resizebox{0.5\textwidth}{!}{
    \begin{tabular}{ccccc}
	\toprule[1.5pt]
	{Method} & {Pretrained Dataset} & {Precision} & {Recall} & {F1} \\
        \hline\hline
        ResNet50 \cite{he2016deep} & ImageNet-1k & \textbf{65.42} & 38.86 & 48.10 \\
        SeCo \cite{manas2021seasonal} & SeCo & 57.71 & 49.23 & 49.82 \\
        ViT \cite{dosovitskiy2020image}& Random Init. & 56.71 & 47.52 &51.71 \\
        ViT-22k \cite{dosovitskiy2020image} & ImageNet-22k & 52.09 & 52.37 & 52.23 \\
        SatMAE \cite{cong2022satmae} & fMoW-S2 & 55.18 & 50.54 & 52.76 \\   
        \hline
        SpectralGPT & fMoW-S2 & 51.65 & 56.15 & 53.51 \\
        SpectralGPT$^{+}$ & fMoW-S2+BigEarthNet & 52.39 & \textbf{57.20} & \textbf{54.29}\\
	\bottomrule[1.5pt]
	\end{tabular}}
	\label{changedetection}
\end{table}

\subsection{Ablation Studies} \label{Ablation Studies}
During the pretraining stage, we conduct a comprehensive study of various factors that may impact downstream task performance. These factors encompass masking ratio, ViT patch size, data scale, reconstruction target, decoder depth, and model size. To provide a more rigorous assessment of pretrained models, we subject all ablation models to fine-tuning on the BigEarthNet multi-label classification dataset with only a 10\% subset of the train set, which poses a more formidable challenge, evaluated using the mAP measurement. Our choice of ViT-B as the backbone model ensures consistency across experiments. Except for ablations involving data scale and training schedule length, all models undergo pretraining on the fMoW-S2 dataset for a duration of 200 epochs. This comprehensive evaluation framework enables us to gain deeper insights into the impact of these factors on model performance.

\begin{figure*}[!t]
      \centering
      \includegraphics[width=1\textwidth]{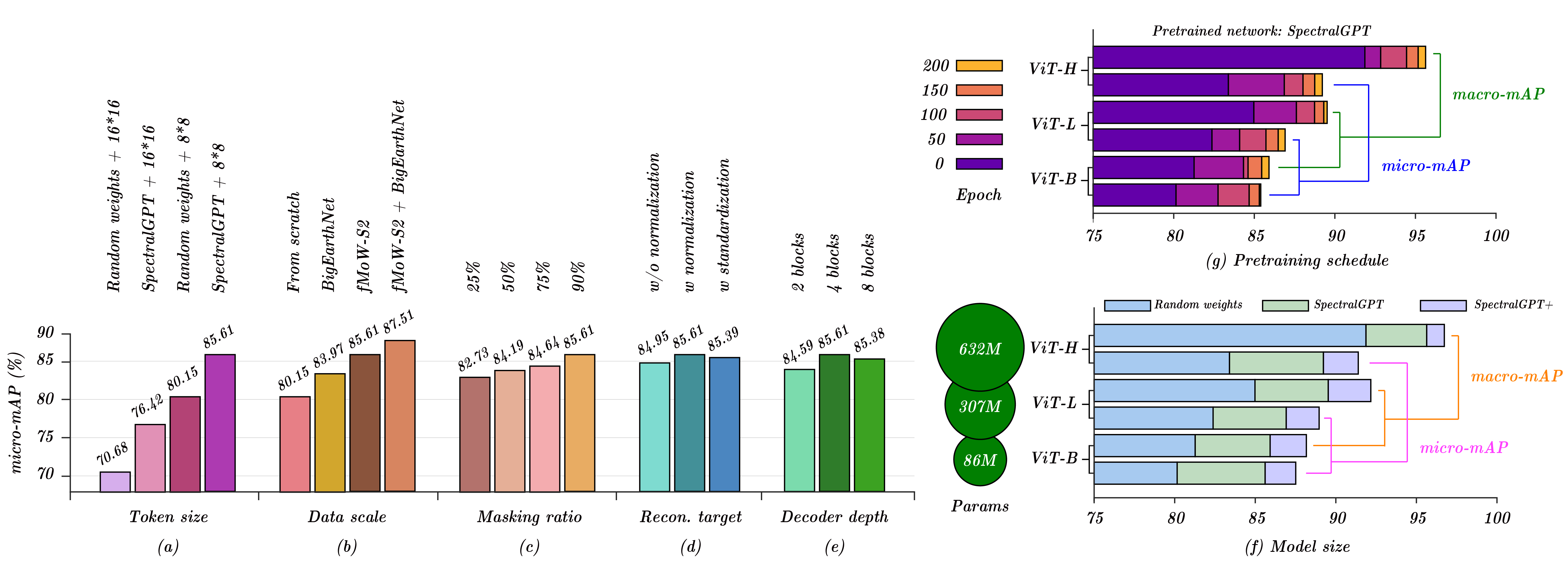}
      \caption{An illustration for ablation analysis of the proposed SpectralGPT foundation model on BigEarthNet-S2 dataset in terms of (a) token size, (b) data scale, (c) masking ratio, (d) reconstruction target, and (e) decoder depth, as well as (f) model size (i.e., ViT-Base, ViT-Large, and ViT-Huge) and their (g) pretraining length/epoch.}
\label{fig:ablation}
\end{figure*}

\begin{table*}[!t]
\caption{Ablation Analysis of the proposed SpectralGPT foundation model in terms of masking ratio, patch size, data scale, reconstruction target, and decoder depth, respectively. The best result is shown in bold.}
\centering 
\subtable[Token Size]{        
  {\begin{tabular}{ccc}
	\toprule
	{Init. Weights} & {Patch Size} & {mAP} \\        
        \hline\hline        
         {Random} & {16} & {70.68} \\        
         {SpectralGPT} & {16} & {76.42} \\        
         {Random} & {8} & {80.15} \\    
         {SpectralGPT} & \textbf{8} & \textbf{85.61}\\
	\bottomrule
	\end{tabular}}
\label{sub_patchsize}
}
\hfill
\subtable[Data Scale]{        
{\begin{tabular}{ccc}
	\toprule
	{Pretrained Dataset} & {mAP} \\        
        \hline\hline        
        {From scratch} & {80.15} \\
        {BigEarthNet} & {83.97} \\
        {fMoW-S2} & {85.61} \\
        {fMoW-S2+BigEarthNet} & \textbf{87.51}\\
	\bottomrule
	\end{tabular}}
\label{sub_datascale}
}
\hfill
\subtable[Masking Ratio]{
\begin{tabular}{cc}
        \toprule
	{Ratio} & {mAP} \\
        \hline\hline
        25$\%$ & 82.73 \\        
        50$\%$ & 84.19 \\        
        75$\%$ & 84.64 \\
        \textbf{90$\%$} & \textbf{85.61} \\  
	\bottomrule
\end{tabular}
\label{sub_maskingratio}
}
\hfill
\subtable[Reconstruction Target]{        
\begin{tabular}{cc}   
	\toprule
	{Case} & {mAP}\\
        \hline\hline
        Without norm & 84.95\\ 
        Normalization & \textbf{85.61}\\
        Standardization & 85.39\\
	\bottomrule	
\end{tabular}
\label{sub_reconstrurtion}
}
\hfill
\subtable[Decoder Depth]{        
  {\begin{tabular}{cc}
	\toprule
	{Blocks} & {mAP}\\
        \hline\hline
        2 & 84.59 \\
        \textbf{4} & \textbf{85.61}\\
        8 & 85.38\\
        \bottomrule
	\end{tabular}}
\label{sub_decoderdepth}
}
\end{table*}

\subsubsection{Token Size}
Table \ref{sub_patchsize} Fig. \ref{fig:ablation}(a) offers crucial insights into the impact of token size on model performance, consistently demonstrating that larger patch sizes lead to reduced model performance, aligning with prior research findings \cite{cong2022satmae}. This phenomenon can be attributed to the intrinsic characteristics of ViT architectures. With larger token sizes, such as $16\times 16$, each image contains fewer tokens, resulting in a diminished level of fine-grained spatial information as the model progresses through its deeper layers. Consequently, this reduction in spatial detail negatively affects the model's overall performance. However, it is noteworthy that the pretrained model consistently enhances mAP regardless of the token size settings, emphasizing its capacity to improve performance across various configurations. Significantly, the recognition performance with a token size of $8\times 8$ is notably superior to that with $16\times 16$, despite the input images being $96\times 96$ or $128\times 128$ in size, underscoring the versatility and efficacy of the pretrained model.

\subsubsection{Data Scale}

Table \ref{sub_datascale} and Fig. \ref{fig:ablation}(b) present a comprehensive analysis focusing on the impact of pretraining data in our research. We conducted pretraining using two datasets (i.e., fMoW-S2, BigEarthNet) while maintaining a standardized input image size of 96$\times$96. To delve deeper into this comparison, we initially pretrained models exclusively on fMoW-S2, followed by a seamless continuation of pretraining on BigEarthNet without any intermediate fine-tuning steps. Our pretraining datasets consisted of the extensive train set of fMoW-S2, which encompasses an impressive 712,874 images from around the world, and the BigEarthNet training set, which comprises 351,496 images within the European region after excluding those affected by snow, clouds, or cloud shadows.

This analysis in Table \ref{sub_datascale} underscores the substantial influence of both data scale and distribution on model pretraining. Models pretrained on the same dataset as the downstream task consistently demonstrate superior performance, highlighting the crucial role of dataset coherence in effective transfer learning. Furthermore, fMoW-S2 outperforms BigEarthNet in pretraining, primarily due to its larger dataset and broader geographic coverage. Interestingly, the concept of continual pretraining, which combines both datasets, results in models with higher mAP scores. This improvement can be attributed in part to the transition from $96\times 96$ images during fMoW-S2 pretraining to $128\times 128$ images during BigEarthNet pretraining, underscoring the beneficial impact of increasing image size on overall model efficacy.

\subsubsection{Masking Ratio}
Table \ref{sub_maskingratio} and  Fig. \ref{fig:ablation}(c) shed light on the impact of the masking ratio, revealing a noteworthy trend where higher masking ratios correspond to improved model performance. Unlike the conventional masking ratio of 75\% often applied to natural RGB images, we find that the optimal masking ratio for multi-spectral images is 90\%. This observation aligns with the hypothesis presented in \cite{feichtenhofer2022masked} that the masking ratio in MIM methods is intricately linked to the information redundancy within the data. Multi-spectral images inherently exhibit greater information redundancy, with strong correlations among their spectral bands. Consequently, a higher masking ratio is essential for the model to effectively learn meaningful representations from these images. Moreover, a 90\% masking ratio significantly enhances the efficiency of the pretraining stage, reducing memory complexity and expediting training times, offering a practical advantage in model development.

\subsubsection{Reconstruction Target}
Table \ref{sub_reconstrurtion} and Fig. \ref{fig:ablation}(d) conduct an insightful analysis of the influence of reconstruction targets on normalized, standardized data and raw data without normalization or standardization in the context of multi-spectral images. Normalization, which scales all data to the [0, 1] range, and standardization, which transforms data to have a mean of 0 and a standard deviation of 1, are the two examined targets. Remarkably, the results show minimal disparity in model performance between normalization and standardization reconstruction targets, primarily because both targets pertain to pixel-level data transformations. However, the model pretrained on raw data performs much worse than the models with normalized reconstruction targets. We attribute this phenomenon to the characteristics of multi-spectral images. The spectral values are usually numerically large and vary from band to band, thereby the model pretraining on the raw data may need a longer pretaining schedule to converge and show the same performance compared with those models pretrained on normalized and standardized data. Our perspective suggests that employing a more semantically meaningful target in a specific representation space could potentially yield improved model performance.

\subsubsection{Decoder Depth}
Table \ref{sub_decoderdepth} and Fig. \ref{fig:ablation}(e) examine the impact of decoder depth on model performance, following the principles of MIM methods where the pretrained encoder serves as the backbone for downstream tasks while discarding the decoder component. Notably, the results reveal that a shallow decoder configuration is ill-suited for spectral model pretraining. This observation aligns with the hypothesis that spectral images, characterized by high dimensionality and complexity, require a decoder with enhanced capacity, consistent with prior findings in the field \cite{feichtenhofer2022masked}.

\subsubsection{Model Size}

Table \ref{modelscale} and Fig. \ref{fig:ablation}(f) give a comparative analysis between fine-tuning results of ViT-B and ViT-L quantitatively and qualitatively, revealing compelling insights. The macro average precision and micro average precision are listed to comprehensively evaluate the performance of models. ViT-B, equipped with 12 transformer layers and 86 million parameters, exhibits promising performance gains when employing the proposed method, achieving a mAP(micro) of 85.41, surpassing the ViT-B trained from scratch by 5.26. On the other hand, ViT-L, featuring 24 layers and 307 million parameters, notably outperforms ViT-B, with a mAP(micro) of 86.92, surpassing the model trained from scratch by a significant margin of 4.44. Besides, ViT-H, concluding 32 layers and 632 million parameters, highly enhances the performance of the neural network on BigEarthNet with the mAP(micro) of 89.23.  Notably, though our models are only fine-tuned with 10\% downstream training data, the ViT-H model employing SpectralGPT$^{+}$ pretrained weights beats all the models even trained with the whole train set, with a SOTA mAP(micro) of 91.39. These results underscore the pivotal role of an appropriate pretraining strategy and indicate that larger ViT models are capable of learning more intricate image representations, rendering them highly suitable for tasks demanding superior accuracy.

\begin{table}[!t]
    \begin{center}
	\caption{Performance comparison using different pretrained models across three ViT-based network scales (i.e., base, large, huge) on the BigEarthNet dataset. The best result is shown in bold.}
     \resizebox{0.5\textwidth}{!}{
	\begin{tabular}{ccccc}
	\toprule[1.5pt]
	{Network Scale} & Params & {Pretained Network} & {macro-mAP} & {micro-mAP} \\
        \hline\hline
        \multirow{3}{*}{ViT-Base} &  \multirow{3}{*}{86M} & Random Init. & 81.27 & 80.15\\  
        & & SpectralGPT & 85.92 & 85.61\\
        & & {SpectralGPT$^{+}$} & \textbf{88.17}  & \textbf{87.50}\\
        \hline
        \multirow{3}{*}{ViT-Large} & \multirow{3}{*}{307M} & Random Init. & 84.98 & 82.38\\
        & & SpectralGPT & 89.53 & 86.92\\
        & & {SpectralGPT$^{+}$} & \textbf{92.17} & \textbf{88.96}\\
        \hline
        \multirow{3}{*}{ViT-Huge} & \multirow{3}{*}{632M} &Random Init.& 91.87 & 83.40\\
        & &SpectralGPT & 95.64  & 89.23\\
        & & {SpectralGPT$^{+}$} & \textbf{96.73} & \textbf{91.39}\\
	\bottomrule[1.5pt]
	\end{tabular}}
	\label{modelscale}
	\end{center}
\end{table}

\subsubsection{Pretraining Schedule}
In Fig. \ref{fig:ablation}(g), we present the fine-tuning results for models trained with varying pre-training epochs, evaluated using the macro-mAP and micro-mAP metrics, respectively. Notably, the models pretrained for just 50 epochs exhibit significant performance gains compared to those trained from scratch. The observed trend in the figure indicates that the models continue to benefit from longer pretraining epochs, suggesting that extended training can further enhance performance. Moreover, the results in Table \ref{modelscale} reinforce this finding, as ViT-L and ViT-H consistently achieve higher mAP compared to ViT-B, highlighting the effectiveness of both extended pretraining and larger model architectures.

\begin{figure*}[!t]
      \centering
      \includegraphics[width=1\textwidth]{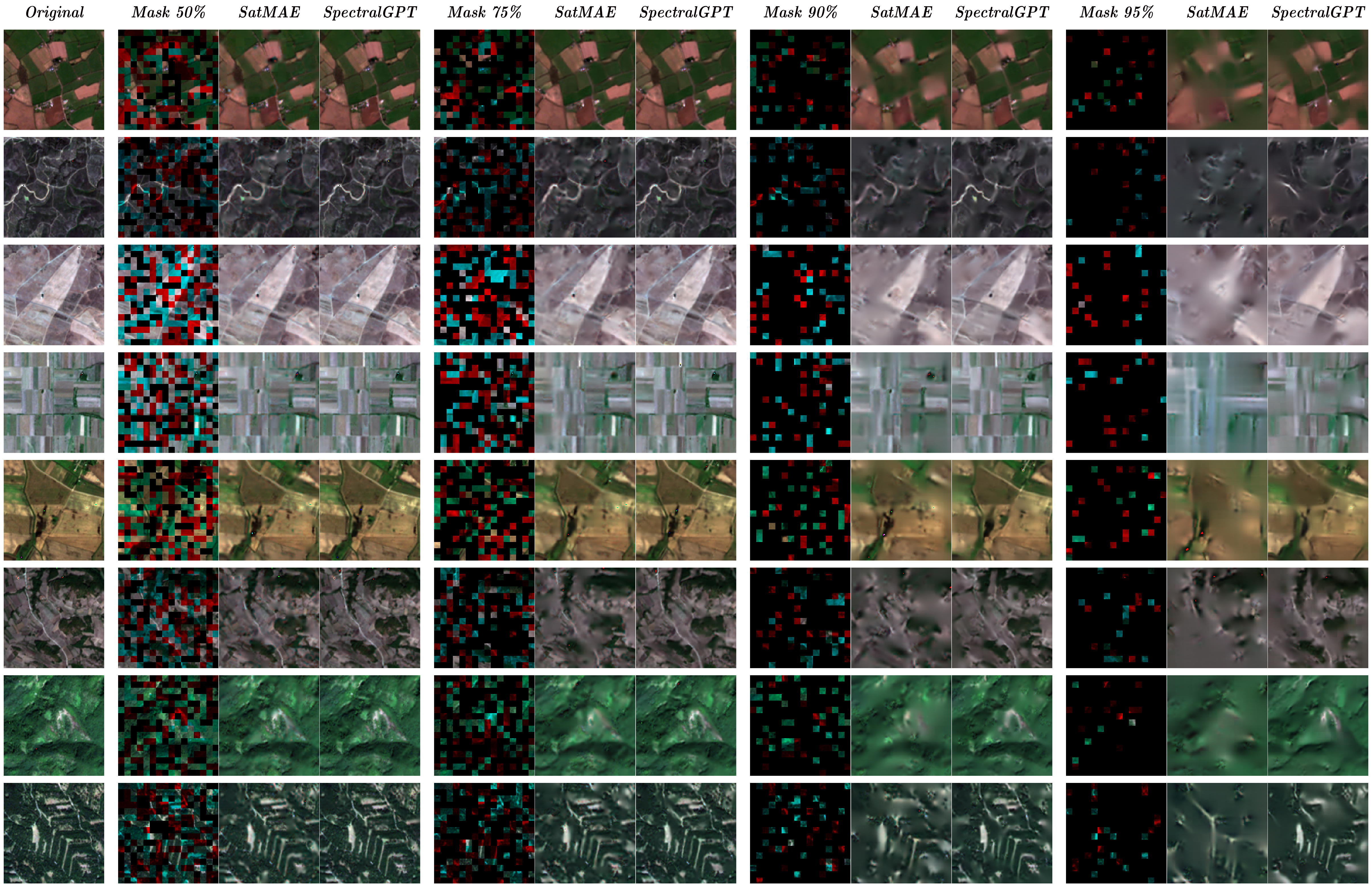}
      \caption{Visual comparison from the nature-color (R: 4, G: 3, B: 2) image reconstruction perspective between SatMAE and SpectralGPT with varied masking ratios of 50\%, 75\%, 90\%, and 95\%, respectively. By masking out a greater number of patches, the reconstructed images exhibit noticeable differences from the originals (e.g., 50\% vs. 95\%), which is expected. It is worth noting, however, that SpectralGPT holds stronger reconstruction capability (\textit{cf. SatMAE}), even if the masking rate has reached over 90\%, showing its powerful learning, reasoning, and generalizing performance.}
\label{fig:reconVis}
\end{figure*}

\begin{figure*}[!t]
      \centering
      \includegraphics[width=1\textwidth]{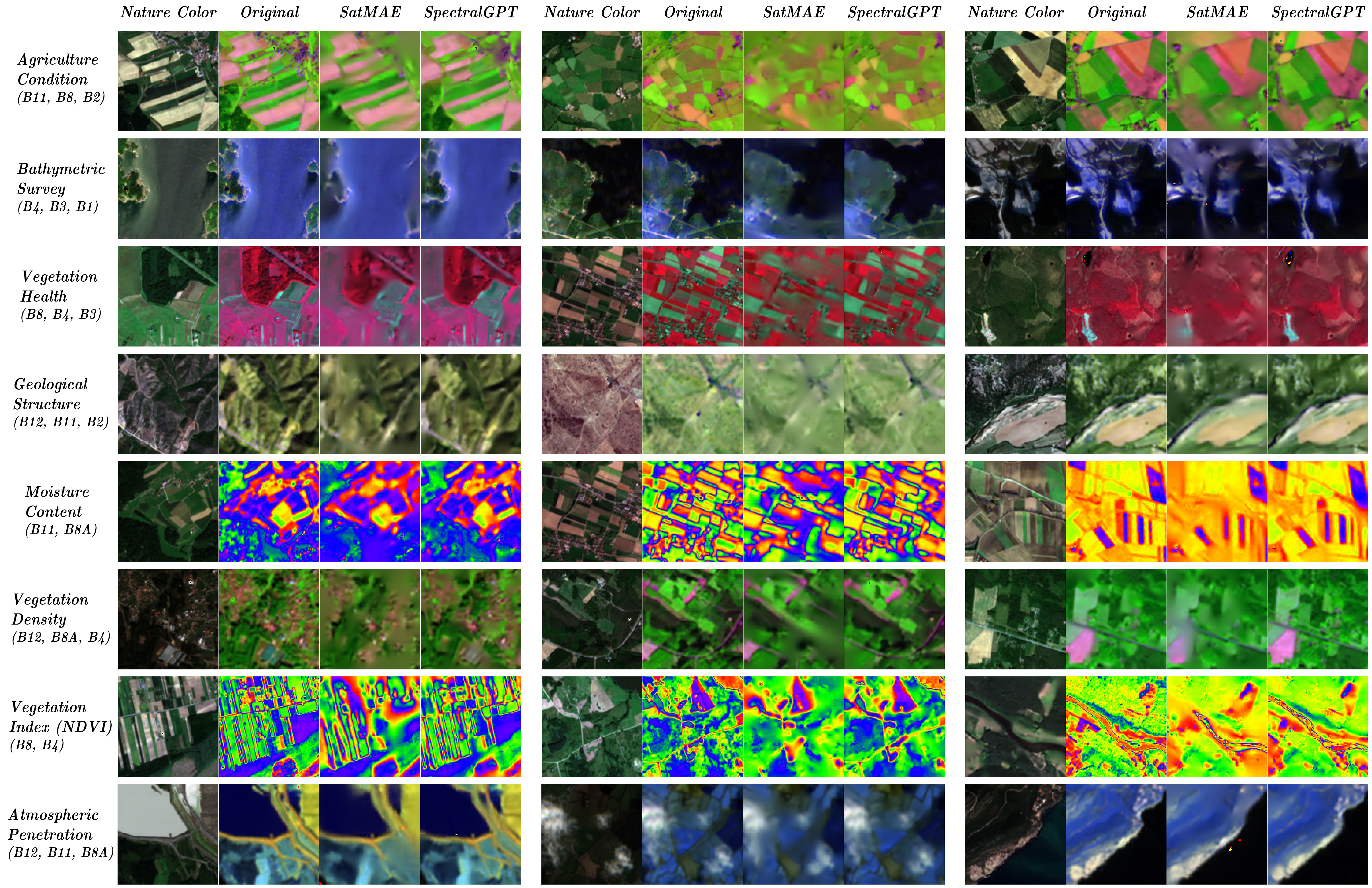}
      \caption{Band combination visualization for geographical characteristic representations of reconstructed spectral images. These images are reconstructed and inferred by SatMAE and SpectralGPT models, respectively, only using 10\% visible patches (masking 90\% patches). Eight band combinations are given to highlight the characteristics, encompassing applications in agriculture, oceanography, geology, vegetation, atmosphere, and more, further serving a variety of EO applications.}
\label{fig:GeoVis}
\end{figure*}

\subsection{Visual Comparison and Geo-characteristic Recoverability}
With varying masking ratios (i.e., 50\%, 75\%, 90\%, and 95\%) as the input, Fig. \ref{fig:reconVis} visually illustrates the image reconstruction results obtained using SatMAE and our SpectralGPT. Not unexpectedly, as the masking ratio increases, the reconstructed images deviate more from the originals. It is worth highlighting, however, that the proposed SpectralGPT outperforms SatMAE significantly in terms of spectral image reconstruction performance, particularly in preserving visual structures and textural details. To be specific, when utilizing 50\% visible patches, the reconstructed results with SatMAE are comparable to those with SpectralGPT, albeit with some slight blurriness in certain fine details in the SatMAE results. As the percentage of masked patches increases (e.g., from a 75\% mask to 90\% and further to 95\%), the reconstruction performance of SatMAE experiences a substantial decline. In contrast, our SpectralGPT exhibits a superior reconstruction capability (\textit{cf.} SatMAE). Even with a masking rate exceeding 90\%, critical structures, and shape components remain preserved in the visuals, demonstrating our model's robust learning, reasoning, and generalization capabilities.

In addition to the in-depth discussion and sensitivity analysis concerning the masking ratio, we have undertaken more extensive investigations into spectral-wise reconstruction capabilities by using only 10\% of visible patches, with the remainder masked out. These investigations prioritize the representation of geographical characteristics, utilizing various spectral band combinations. As illustrated in Fig. \ref{fig:GeoVis}, we present visualizations of eight different band combinations. These visualizations distinctly highlight the remarkable superiority of our proposed SpectralGPT (closer to that generated by original images) when compared to SatMAE, particularly in terms of band-wise spectral reconstruction capabilities and its application value in the context of EO tasks. In our study, eight geo-characteristics have been identified using various band combinations, corresponding to observation targets in practical applications\footnote{https://www.sentinel-hub.com/}, as detailed in Table \ref{tab:Geo-chara}. Furthermore, notable visual differences are evident in the geo-characteristics obtained using SatMAE and SpectralGPT. These pronounced visual disparities can be attributed to spectral degradation stemming from the comparatively limited reconstructive and inferential capabilities of SatMAE when compared to our more powerful SpectralGPT.

\begin{table}[!t]
    \begin{center}
	\caption{List of geographical characteristics and visualized band combinations corresponding to observation targets in practical geoscience applications.}
    \resizebox{0.5\textwidth}{!}{
	\begin{tabular}{ccc}
	\toprule[1.5pt]
	{Geo-characteristic} & {Band Combination} & {Observation Target} \\
        \hline        
        Agriculture Condition & B11, B8, B2 &  Crop Health\\
        Bathymetric Survey & B4, B3, B1 & Coast\\
        Vegetation Health & B8, B4, B3 &  Chlorophyll \\ 
        Geological Structure & B12, B11, B2 & Faults, Lithology\\
        Moisture Content & B11, B8A & Plant Water Pressure\\
        Vegetation Density & B12, B8A, B4 & Vegetation Cover, Soil, Building\\
        Vegetation Index (NDVI) & B8, B4 & Tree Crown, Urban, Waterscape \\
        Atmospheric Penetration & B12, B11, B8A & Particle, Smoke, Haze, Thin Cloud\\
	\bottomrule[1.5pt]
	\end{tabular}
     }
	\label{tab:Geo-chara}
	\end{center}
\end{table}

\section{Conclusion}
The explosive development of foundation models represents a significant technological revolution following the advent of deep learning. Currently, various industries are witnessing significant leaps in technology and application advancements, largely driven by the emergence of foundation models. The RS field is no exception, with numerous EO applications, reaping significant benefits. 

Spectral imaging has gained recognition in EO  for its ability to provide rich insights into the composition of observed objects and materials, making it a transformative technology with vast potential to address global challenges and reshape various industries. However, the ever-expanding availability of spectral data from various RS platforms undeniably presents formidable challenges. There is a pressing demand for the development of foundation models specifically designed for spectral RS data. To fully unlock and leverage the potential of spectral RS data, it is imperative to overcome and resolve several challenging obstacles. These include efficiently processing and utilizing diverse RS spectral big data from various sources, extracting meaningful knowledge representations from complex spatial-spectral mixed information, and tackling the spectral degradation of neighboring spectral relevance modeling.

In response to these challenges, we propose SpectralGPT, a customized spectral RS foundation model, featuring a novel 3D GPT architecture. With its innovative 3D GPT architecture, trained on over one million spectral images and over 600 million parameters, SpectralGPT empowers intelligent processing of spectral RS big data. SpectralGPT can flexibly handle diverse inputs in terms of size, resolution, temporal variability, and geographical coverage. This 3D masking strategy enables effective information extraction from spatial-spectral coupling tokens. Moreover, the innovative multi-target reconstruction is capable of capturing sequentially preserving spectral characteristics and meanwhile mitigating spectral degradation. Notably, our progressive training paradigm empowers the foundation model, surpassing transitional points in performance. These breakthroughs achieved by SpectralGPT democratize access to spectral RS big data, rendering it more accessible and cost-effective for large-scale EO applications.

Our study also includes a comprehensive assessment of MAE-based pretrained foundation models, with a focus on spectral reconstruction capabilities. We systematically evaluated model performance with inputs ranging from 50\% to as low as 5\% visible tokens. This extensive analysis allows us to gauge their proficiency in spectral-wise reconstruction and inference, especially significant in Geo-fields, such as agriculture, oceanography, geology, and vegetation. Visualizing the band combinations of reconstructed spectral images using both SatMAE and SpectralGPT demonstrates the latter's potential in practical EO tasks and Geo-field applications.

Looking ahead, our research will pursue several objectives. We plan to expand the volume and diversity of RS data used for training, encompassing various modalities, resolutions, time series, and image sizes. This enrichment will enhance the robustness of the RS foundation model. Furthermore, we aim to extend SpectralGPT's capabilities by incorporating a wider range of downstream tasks. This will transform SpectralGPT into a versatile AI model with improved generalization, well-suited for diverse EO and geoscience applications.

\ifCLASSOPTIONcompsoc
  \section*{Acknowledgments}
\else
  \section*{Acknowledgment}
\fi

This work was supported by the National Natural Science Foundation of China under Grant 42030111, 42271350, 42241109, and 62201553, and the MIAI@Grenoble Alpes (ANR-19-P3IA-0003).

\bibliographystyle{ieeetr}
\bibliography{HDF_ref}

\begin{thebibliography}{10}

\bibitem{goetz1985imaging}
A.~F. Goetz, G.~Vane, J.~E. Solomon, and B.~N. Rock, ``Imaging spectrometry for earth remote sensing,'' {\em Science}, vol.~228, no.~4704, pp.~1147--1153, 1985.

\bibitem{mei2022hyperspectral}
S.~Mei, C.~Song, M.~Ma, and F.~Xu, ``Hyperspectral image classification using group-aware hierarchical transformer,'' {\em IEEE Transactions on Geoscience and Remote Sensing}, vol.~60, pp.~1--14, 2022.

\bibitem{zhou2023rethinking}
M.~Zhou, J.~Huang, D.~Hong, F.~Zhao, C.~Li, and J.~Chanussot, ``Rethinking pan-sharpening in closed-loop regularization,'' {\em IEEE Transactions on Neural Networks and Learning Systems}, 2023.

\bibitem{reichstein2019deep}
M.~Reichstein, G.~Camps-Valls, B.~Stevens, M.~Jung, J.~Denzler, N.~Carvalhais, and f.~Prabhat, ``Deep learning and process understanding for data-driven earth system science,'' {\em Nature}, vol.~566, no.~7743, pp.~195--204, 2019.

\bibitem{he2020non}
W.~He, Q.~Yao, C.~Li, N.~Yokoya, Q.~Zhao, H.~Zhang, and L.~Zhang, ``Non-local meets global: An iterative paradigm for hyperspectral image restoration,'' {\em IEEE Transactions on Pattern Analysis and Machine Intelligence}, vol.~44, no.~4, pp.~2089--2107, 2020.

\bibitem{hong2023decoupled}
D.~Hong, J.~Yao, C.~Li, D.~Meng, N.~Yokoya, and J.~Chanussot, ``Decoupled-and-coupled networks: Self-supervised hyperspectral image super-resolution with subpixel fusion,'' {\em IEEE Transactions on Geoscience and Remote Sensing}, 2023.

\bibitem{mei2023rotation}
S.~Mei, R.~Jiang, M.~Ma, and C.~Song, ``Rotation-invariant feature learning via convolutional neural network with cyclic polar coordinates convolutional layer,'' {\em IEEE Transactions on Geoscience and Remote Sensing}, vol.~61, pp.~1--13, 2023.

\bibitem{hong2023cross}
D.~Hong, B.~Zhang, H.~Li, Y.~Li, J.~Yao, C.~Li, M.~Werner, J.~Chanussot, A.~Zipf, and X.~X. Zhu, ``Cross-city matters: A multimodal remote sensing benchmark dataset for cross-city semantic segmentation using high-resolution domain adaptation networks,'' {\em Remote Sensing of Environment}, vol.~299, p.~113856, 2023.

\bibitem{bommasani2021opportunities}
R.~Bommasani, D.~A. Hudson, E.~Adeli, R.~Altman, S.~Arora, S.~von Arx, M.~S. Bernstein, J.~Bohg, A.~Bosselut, E.~Brunskill, {\em et~al.}, ``On the opportunities and risks of foundation models,'' {\em arXiv preprint arXiv:2108.07258}, 2021.

\bibitem{tian2023recent}
J.~Tian, X.~Sun, Y.~Du, S.~Zhao, Q.~Liu, K.~Zhang, W.~Yi, W.~Huang, C.~Wang, X.~Wu, {\em et~al.}, ``Recent advances for quantum neural networks in generative learning,'' {\em IEEE Transactions on Pattern Analysis and Machine Intelligence}, 2023.

\bibitem{vaswani2017attention}
A.~Vaswani, N.~Shazeer, N.~Parmar, J.~Uszkoreit, L.~Jones, A.~N. Gomez, {\L}.~Kaiser, and I.~Polosukhin, ``Attention is all you need,'' {\em NeurIPS}, vol.~30, 2017.

\bibitem{radford2021learning}
A.~Radford, J.~W. Kim, C.~Hallacy, A.~Ramesh, G.~Goh, S.~Agarwal, G.~Sastry, A.~Askell, P.~Mishkin, J.~Clark, {\em et~al.}, ``Learning transferable visual models from natural language supervision,'' in {\em ICML}, pp.~8748--8763, PMLR, 2021.

\bibitem{liu2021modality}
X.~Liu, D.~Hong, J.~Chanussot, B.~Zhao, and P.~Ghamisi, ``Modality translation in remote sensing time series,'' {\em IEEE Transactions on Geoscience and Remote Sensing}, vol.~60, pp.~1--14, 2021.

\bibitem{he2020momentum}
K.~He, H.~Fan, Y.~Wu, S.~Xie, and R.~Girshick, ``Momentum contrast for unsupervised visual representation learning,'' in {\em CVPR}, pp.~9729--9738, 2020.

\bibitem{chen2020simple}
T.~Chen, S.~Kornblith, M.~Norouzi, and G.~Hinton, ``A simple framework for contrastive learning of visual representations,'' in {\em ICML}, pp.~1597--1607, 2020.

\bibitem{chen2020big}
T.~Chen, S.~Kornblith, K.~Swersky, M.~Norouzi, and G.~E. Hinton, ``Big self-supervised models are strong semi-supervised learners,'' {\em NeurIPS}, vol.~33, pp.~22243--22255, 2020.

\bibitem{caron2020unsupervised}
M.~Caron, I.~Misra, J.~Mairal, P.~Goyal, P.~Bojanowski, and A.~Joulin, ``Unsupervised learning of visual features by contrasting cluster assignments,'' {\em NeurIPS}, vol.~33, pp.~9912--9924, 2020.

\bibitem{chen2021exploring}
X.~Chen and K.~He, ``Exploring simple siamese representation learning,'' in {\em CVPR}, pp.~15750--15758, 2021.

\bibitem{chen2020improved}
X.~Chen, H.~Fan, R.~Girshick, and K.~He, ``Improved baselines with momentum contrastive learning,'' {\em arXiv preprint arXiv:2003.04297}, 2020.

\bibitem{xiong2020loco}
Y.~Xiong, M.~Ren, and R.~Urtasun, ``Loco: Local contrastive representation learning,'' {\em NeurIPS}, vol.~33, pp.~11142--11153, 2020.

\bibitem{xie2021propagate}
Z.~Xie, Y.~Lin, Z.~Zhang, Y.~Cao, S.~Lin, and H.~Hu, ``Propagate yourself: Exploring pixel-level consistency for unsupervised visual representation learning,'' in {\em CVPR}, pp.~16684--16693, 2021.

\bibitem{dosovitskiy2020image}
A.~Dosovitskiy, L.~Beyer, A.~Kolesnikov, D.~Weissenborn, X.~Zhai, T.~Unterthiner, M.~Dehghani, M.~Minderer, G.~Heigold, S.~Gelly, {\em et~al.}, ``An image is worth 16x16 words: Transformers for image recognition at scale,'' {\em arXiv preprint arXiv:2010.11929}, 2020.

\bibitem{bao2022beit}
H.~Bao, L.~Dong, S.~Piao, and F.~Wei, ``Beit: Bert pre-training of image transformers,'' in {\em ICLR}, 2022.

\bibitem{du2023spectral}
D.~Du, Y.~Gu, T.~Liu, and X.~Li, ``Spectral reconstruction from satellite multispectral imagery using convolution and transformer joint network,'' {\em IEEE Transactions on Geoscience and Remote Sensing}, 2023.

\bibitem{he2022masked}
K.~He, X.~Chen, S.~Xie, Y.~Li, P.~Doll{\'a}r, and R.~Girshick, ``Masked autoencoders are scalable vision learners,'' in {\em CVPR}, pp.~16000--16009, 2022.

\bibitem{wang2022advancing}
D.~Wang, Q.~Zhang, Y.~Xu, J.~Zhang, B.~Du, D.~Tao, and L.~Zhang, ``Advancing plain vision transformer towards remote sensing foundation model,'' {\em IEEE Transactions on Geoscience and Remote Sensing}, 2022.

\bibitem{sun2022ringmo}
X.~Sun, P.~Wang, W.~Lu, Z.~Zhu, X.~Lu, Q.~He, J.~Li, X.~Rong, Z.~Yang, H.~Chang, {\em et~al.}, ``Ringmo: A remote sensing foundation model with masked image modeling,'' {\em IEEE Transactions on Geoscience and Remote Sensing}, 2022.

\bibitem{liu2021swin}
Z.~Liu, Y.~Lin, Y.~Cao, H.~Hu, Y.~Wei, Z.~Zhang, S.~Lin, and B.~Guo, ``Swin transformer: Hierarchical vision transformer using shifted windows,'' in {\em ICCV}, pp.~10012--10022, 2021.

\bibitem{tong2022videomae}
Z.~Tong, Y.~Song, J.~Wang, and L.~Wang, ``Videomae: Masked autoencoders are data-efficient learners for self-supervised video pre-training,'' {\em NeurIPS}, vol.~35, pp.~10078--10093, 2022.

\bibitem{feichtenhofer2022masked}
C.~Feichtenhofer, Y.~Li, K.~He, {\em et~al.}, ``Masked autoencoders as spatiotemporal learners,'' {\em NeurIPS}, vol.~35, pp.~35946--35958, 2022.

\bibitem{cong2022satmae}
Y.~Cong, S.~Khanna, C.~Meng, P.~Liu, E.~Rozi, Y.~He, M.~Burke, D.~Lobell, and S.~Ermon, ``Satmae: Pre-training transformers for temporal and multi-spectral satellite imagery,'' {\em NeurIPS}, vol.~35, pp.~197--211, 2022.

\bibitem{vincent2010stacked}
P.~Vincent, H.~Larochelle, I.~Lajoie, Y.~Bengio, P.-A. Manzagol, and L.~Bottou, ``Stacked denoising autoencoders: Learning useful representations in a deep network with a local denoising criterion.,'' {\em Journal of Machine Learning Research}, vol.~11, no.~12, 2010.

\bibitem{devlin2018bert}
J.~Devlin, M.-W. Chang, K.~Lee, and K.~Toutanova, ``Bert: Pre-training of deep bidirectional transformers for language understanding,'' {\em arXiv preprint arXiv:1810.04805}, 2018.

\bibitem{christie2018functional}
G.~Christie, N.~Fendley, J.~Wilson, and R.~Mukherjee, ``Functional map of the world,'' in {\em CVPR}, pp.~6172--6180, 2018.

\bibitem{sumbul2019bigearthnet}
G.~Sumbul, M.~Charfuelan, B.~Demir, and V.~Markl, ``Bigearthnet: A large-scale benchmark archive for remote sensing image understanding,'' in {\em IGARSS}, pp.~5901--5904, IEEE, 2019.

\bibitem{loshchilov2019decoupled}
I.~Loshchilov and F.~Hutter, ``Decoupled weight decay regularization,'' in {\em ICLR}, 2019.

\bibitem{he2016deep}
K.~He, X.~Zhang, S.~Ren, and J.~Sun, ``Deep residual learning for image recognition,'' in {\em CVPR}, pp.~770--778, 2016.

\bibitem{manas2021seasonal}
O.~Manas, A.~Lacoste, X.~Gir{\'o}-i Nieto, D.~Vazquez, and P.~Rodriguez, ``Seasonal contrast: Unsupervised pre-training from uncurated remote sensing data,'' in {\em ICCV}, pp.~9414--9423, 2021.

\bibitem{helber2019eurosat}
P.~Helber, B.~Bischke, A.~Dengel, and D.~Borth, ``Eurosat: A novel dataset and deep learning benchmark for land use and land cover classification,'' {\em IEEE Journal of Selected Topics in Applied Earth Observations and Remote Sensing}, vol.~12, no.~7, pp.~2217--2226, 2019.

\bibitem{neumann2020domain}
M.~Neumann, A.~S. Pinto, X.~Zhai, and N.~Houlsby, ``In-domain representation learning for remote sensing,'' in {\em ICLR-AI for Earth Sciences Workshop}, 2020.

\bibitem{xiao2018unified}
T.~Xiao, Y.~Liu, B.~Zhou, Y.~Jiang, and J.~Sun, ``Unified perceptual parsing for scene understanding,'' in {\em ECCV}, pp.~418--434, 2018.

\bibitem{hong_danfeng_2023_8412377}
D.~Hong, B.~Zhang, X.~Li, Y.~Li, C.~Li, J.~Yao, N.~Yokoya, H.~Li, X.~Jia, A.~Plaza, P.~Gamba, J.~A. Benediktsson, and J.~Chanussot, ``{SpectralGPT: The first remote sensing foundation model customized for spectral data},'' Oct. 2023.

\bibitem{daudt2018urban}
R.~C. Daudt, B.~Le~Saux, A.~Boulch, and Y.~Gousseau, ``Urban change detection for multispectral earth observation using convolutional neural networks,'' in {\em IGARSS}, pp.~2115--2118, Ieee, 2018.

\end{thebibliography}


\end{document}